\documentclass[journal]{IEEEtran}

\ifCLASSINFOpdf
  \usepackage[pdftex]{graphicx}
\else
\fi

\usepackage[cmex10]{amsmath}
\usepackage{amssymb}
\usepackage{caption}

\ifCLASSOPTIONcompsoc
 \usepackage[caption=false,font=normalsize,labelfont=sf,textfont=sf]{subfig}
\else
 \usepackage[caption=false,font=footnotesize]{subfig}

\usepackage{hyperref}
\usepackage{multirow}
\usepackage{threeparttable}
\usepackage{amsmath,graphicx,bm,threeparttable,indentfirst,cite}
\usepackage{algorithm, algorithmic,booktabs}
\usepackage{color, multirow,graphicx}
\usepackage{multirow,epsfig,fbox,amsfonts,amsmath,multicol,enumitem,color,pifont,booktabs}
\usepackage{hyperref}
\usepackage{bbding}
\hypersetup{
	colorlinks=true,
	linkcolor=blue,
        citecolor=blue
}
\usepackage{pifont}

\begin{document}
%
\title{Semantic-Spatial Feature Fusion with Dynamic Graph Refinement for Remote Sensing Image Captioning}
%
%
%

\author{Maofu~Liu,~\IEEEmembership{Member,~IEEE,}
        Jiahui~Liu,~\IEEEmembership{Member,~IEEE,}
        and~Xiaokang~Zhang,~\IEEEmembership{Senior Member,~IEEE}
\thanks{This work was supported in part by the National Natural Science Foundation of China under Grant No. 42371374 and the "14th Five Year Plan" Advantageous and Characteristic Discipline Project of Hubei Province (China) under Grant No. 2023D0302. (\textit{Corresponding author: Xiaokang Zhang}).}
\thanks{Maofu Liu and Jiahui Liu are with the  School of Computer Science and Technology, Wuhan University of Science and Technology, Wuhan 430065, China. (e-mail: liumaofu@wust.edu.cn; LiuJH@wust.edu.cn).}
\thanks{Xiaokang Zhang is with the School of Information Science and Engineering, Wuhan University of Science and Technology, Wuhan 430081, China (e-mail: natezhangxk@gmail.com).}
}

%
%

\markboth{Journal of \LaTeX\ Class Files,~Vol.~13, No.~9, September~2014}%
{Shell \MakeLowercase{\textit{et al.}}: Bare Demo of IEEEtran.cls for Journals}
%



\maketitle

\begin{abstract}
Remote sensing image captioning aims to generate semantically accurate descriptions that are closely linked to the visual features of remote sensing images. Existing approaches typically emphasize fine-grained extraction of visual features and capturing global information. However, they often overlook the complementary role of textual information in enhancing visual semantics and face challenges in precisely locating objects that are most relevant to the image context. To address these challenges, this paper presents a semantic-spatial feature fusion with dynamic graph refinement (SFDR) method, which integrates the semantic-spatial feature fusion (SSFF) and dynamic graph feature refinement (DGFR) modules. The SSFF module utilizes a multi-level feature representation strategy by leveraging pre-trained CLIP features, grid features, and ROI features to integrate rich semantic and spatial information. In the DGFR module, a graph attention network captures the relationships between feature nodes, while a dynamic weighting mechanism prioritizes objects that are most relevant to the current scene and suppresses less significant ones. Therefore, the proposed SFDR method significantly enhances the quality of the generated descriptions. Experimental results on three benchmark datasets demonstrate the effectiveness of the proposed method. The source code will be available at \href{https://github.com/zxk688}{https://github.com/zxk688}.
\end{abstract}


%
\IEEEpeerreviewmaketitle

\section{Introduction}
%
%
%
%
\IEEEPARstart{T}{he} image captioning task \cite{r1,r2,r3,r4} represents a cross-disciplinary research domain at the intersection of computer vision and natural language processing. Its primary objective is to generate concise and comprehensible sentences closely aligned with image content, thereby bridging the gap between visual data and human understanding. With the rapid advancements in cross-modal research \cite{n10}, image captioning tasks using deep learning methods have gained significant attention in recent years \cite{r5}. The exponential growth of data generated by remote sensing platforms, such as drones and satellites, has progressively extended image captioning to remote sensing image captioning (RSIC) task \cite{r6}. With extensive potential applications in urban planning, military analysis, and environmental monitoring, RSIC has demonstrated remarkable research value and competitiveness. Unlike other tasks, such as object detection \cite{r7,r8}, image classification \cite{r9,r10,r11} or scene classification\cite{n7,n8}, which focus on predicting single labels or keywords, RSIC aims to generate semantically rich and meaningful descriptions for remote sensing images. This distinction highlights RSIC's unique capability to bridge vision and language, enabling a deeper multimodal understanding and representation of remote sensing data.

In the RSIC task, the encoder-decoder architecture has become the dominant framework, gradually replacing traditional template-based \cite{r12} and retrieval-based methods \cite{r13}. This architecture typically leverages convolutional neural networks (CNNs) as encoders to extract features from images, which are subsequently decoded into natural language sentences using decoders such as long short-term memory (LSTM) or Transformer. Existing research primarily focuses on feature extraction and representation. Specifically, VGG and ResNet served as the backbones to extract visual features \cite{r14,r15}. However, the unique imaging characteristics and complex content representations of remote sensing images render single-level feature extraction insufficient to capture the diverse information embedded in these images comprehensively. Therefore, multi-level feature representations \cite{r16,r17} have been explored to address the challenges posed by the characteristics of multi-scale ground objects \cite{r18,r19,r20}. Moreover, attention mechanisms have widely been conducted on visual features to enhance the interpretability and precision of generated descriptions \cite{r16,r21,r22}, while capturing finer details within remote sensing images \cite{r23,r24,r25}. Besides, due to the capability of reducing the modality gap, the visual-language pretraining models (VLPMs) have gained increasing attention in multimodal understanding and generation tasks \cite{n3,n4,n5,n6,r26}.

\begin{figure}[tp]
    \centering
    \includegraphics[width=1\linewidth]{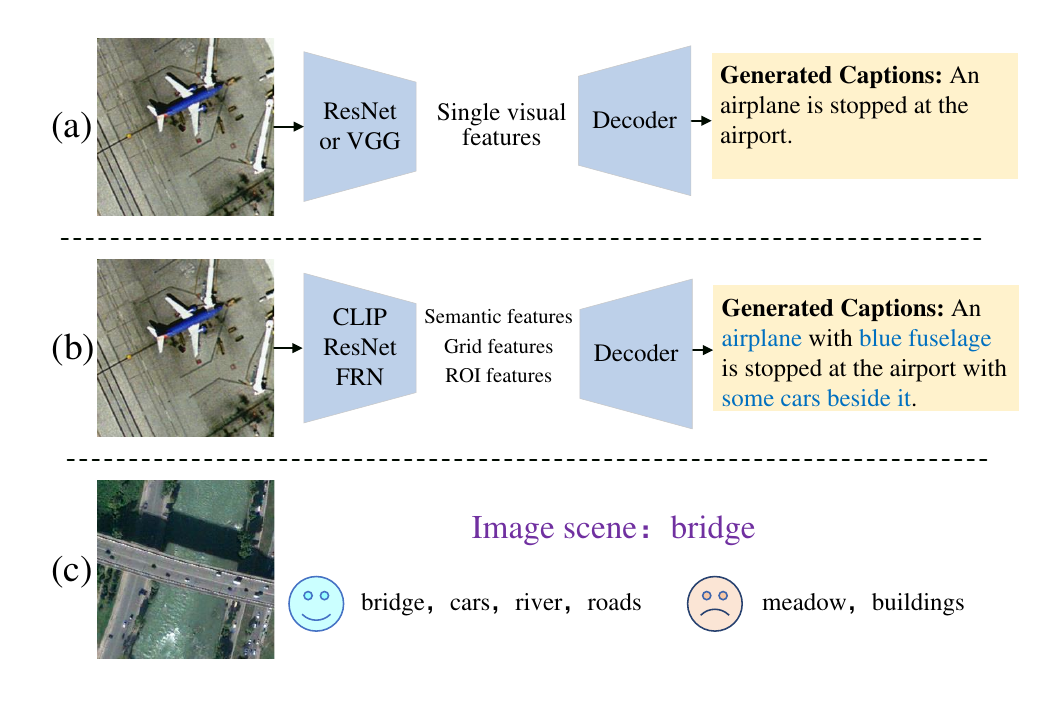}
    \caption{The motivation of our research. (a) Traditional RSIC methods primarily rely on single-level visual features to generate captions. (b) In comparison, we use multi-level feature representations to provide a more comprehensive understanding of semantic information, spatial scenes, and object details. (c) The challenge lies in the interference from unrelated objects in a heterogeneous background, which affects the model’s ability to accurately describe the target regions.}
    \label{fig1}
\end{figure}

Despite notable advancements in RSIC, effectively bridging the modality gap between visual and textual data remains a pressing issue. Specifically, feature representation directly impacts overall captioning performance, as illustrated in Fig. \ref{fig1}. However, a single encoder using ResNet or VGG to extract visual features is limited in capturing the semantics, scenes, and objects of remote sensing images. Additionally, the properties of the large field of view in image acquisition introduce unique challenges, such as covering vast spatial areas and diverse objects, often resulting in misidentification or including irrelevant objects in generated text descriptions. For example, as shown in Fig. \ref{fig1}(c), in a ``bridge'' scene, the captioning model should focus on objects such as ``bridge'', ``cars'', ``river'' and ``road'', while minimizing the influence of irrelevant elements such as ``meadow'' and ``buildings''. Therefore, multi-level feature representations are required to accurately identify scene-relevant objects while suppressing irrelevant information.

To address the aforementioned limitations, we propose a novel RSIC method named semantic-spatial feature fusion with dynamic graph refinement (SFDR), which consists of semantic-spatial feature fusion (SSFF) and dynamic graph feature refinement (DGFR). First, a pre-trained CLIP model is utilized to extract visual and textual features, which can be mapped into a unified embedding space for multimodal feature alignment. Furthermore, we extract grid features and ROI features to enhance the understanding of remote sensing scenes and objects. By constructing multi-level feature representations, the model improves scene comprehension and reduces semantic interpretation bias. To focus on objects most relevant to the current scene, we incorporate a graph attention network (GAT) to model the contextual relationships between object features, which are regarded as nodes in the graph. Additionally, a dynamic weighting mechanism is designed to adaptively adjust attention allocation based on feature importance. As a result, it can emphasize important features while suppressing irrelevant ones. In summary, the main contributions of this paper are as follows:
\begin{enumerate}
    \item The novel SSFF module and DGFR module were proposed to enhance image semantic understanding and object representation capabilities for the RSIC. This has been verified by comparison experiments on three benchmark datasets and $14$ state-of-the-art methods.
    \item We constructed a multi-level feature representation framework that fully leverages CLIP features, grid features, and ROI features to enhance the understanding of semantics, scenes, and objects, effectively mitigating interpretation bias.
    \item We proposed a dynamic graph feature refinement approach. It models feature relationships based on graph attention network and adaptively adjusts feature importance through the proposed dynamic weighting mechanism. This enhances feature representation and effectively guides the description generation process.
\end{enumerate}

This article is structured as follows: Section \ref{sec2} provides a brief review of related research on remote sensing image captioning. Section \ref{sec3} describes the key components and training scheme of the proposed method in detail. Section \ref{sec4} presents the performance evaluations of our method along with comparisons to existing methods. Finally, Section \ref{sec5} concludes the article.

\section{Related Work}\label{sec2}

\subsection{Feature Representation in RSIC}
Effective feature representation of remote sensing images is a critical task for understanding image content. Notably, Qu et al. \cite{r6} leveraged the powerful feature extraction capabilities of CNN by combining CNN image features with the recurrent neural network (RNN) to generate image descriptions. Due to the significant differences between remote sensing images and natural images, multi-level feature extraction is often required to adequately capture the content of remote sensing images. Specifically, Meng et al. \cite{r16} employed a multi-level feature extraction strategy, utilizing ResNet and Faster R-CNN to extract scene-level and object-level features, respectively, thereby bridging the connection between scenes and objects. Furthermore, Zhao et al. \cite{r17} combined grid and region features to leverage fine-grained details and object-level context, addressing their individual limitations. To enhance feature representations, Wang et al. \cite{r29} explored sentence-level and class-level features from different views. These methods address the limitations of single-feature representation but still struggle to fully capture the multi-scale nature of remote sensing scenes. To tackle this, Ma et al. \cite{r18} applied multi-scale attention for hierarchical feature extraction and employed a multi-head attention mechanism to capture contextual information. Liu et al. \cite{r19} incorporated features from different ResNet50 layers into the encoder to enhance multi-scale representation. Wu et al. \cite{r20} utilized a Swin Transformer encoder with a shifted window partitioning scheme to extract multi-scale features while uncovering relationships between objects. Similarly, Yang et al. \cite{r28} introduced the Interactive Fourier Transformer, where a Fourier layer extracts multi-scale features in the frequency domain, effectively reducing redundancy in visual features. Meng et al. \cite{n2}  proposed a multi-scale grouping Transformer with CLIP latents, in which the encoder utilizes dilated convolutions with different dilation rates to obtain multi-scale visual features.
However, these methods primarily employ multi-scale aggregation modules to explore the spatial relationships and multi-scale information of objects within remote sensing images but fail to effectively utilize both spatial information and semantic features.


\subsection{Attention Mechanism in RSIC}
To reduce the semantic gap between features at different levels, Zhang et al. \cite{r30} investigated the influence of attributes extracted from images on the attention mechanism, while Zhang et al. \cite{r31} leveraged word embedding vectors of predicted categories to guide the computation of attention masks. Subsequently, Zhang et al. \cite{r32} developed a global visual feature-guided attention (GVFGA) mechanism, which integrates global visual features with local visual features to establish their relationships. What’s more, to further refine the fusion of features, an LS-guided attention (LSGA) mechanism is proposed. Li et al. \cite{r21} proposed a recursive attention and semantic gate (RASG) framework that combines competitive visual features with a recursive attention mechanism. This method generates improved contextual vectors for the image at each step, thereby enhancing word state representations. Additionally, Zhang et al. \cite{r22} introduced a stair attention mechanism that divides attention weights into three levels—core regions, surrounding regions, and other regions—based on the degree of focus on visual areas, improving the ability to capture relevant visual regions. Li et al. \cite{r35} developed a visual-semantic co-attention (VSCA) mechanism to calculate concept-related regions and region-related concepts, enabling effective multimodal interaction.
Although attention mechanisms can partially localize details and visually relevant regions, they are still weak in capturing complex relationships between scenes and objects, as well as interactions among objects, within the context of remote sensing.

\subsection{Cross-modal Learning in RSIC}
Considering the differences between modalities, Yang et al. \cite{r33} designed a cross-modal feature interaction module in the decoder to promote feature alignment. This module fully utilizes cross-modal features to locate key objects. Li et al. \cite{r34} proposed a novel cross-modal retrieval method that employs a cross-modal retrieval model to retrieve relevant sentences for each image. The words in these retrieved sentences are treated as primary semantic information, providing valuable supplementary information for caption generation. Besides, Li et al. \cite{r35} incorporated the two types of attentive vectors with semantic-level relational features into a consensus exploitation block for learning cross-modal consensus-aware knowledge. Wei et al. \cite{n1} aligned visual and language modalities with a pretrained CLIP model, thus optimizing the domain-specific prompt vector from two aspects: attribute and semantic consistency. Moreover, Yang et al. \cite{r28} designed a two-stage vision–language training method aimed at aligning visual features with textual information. The features of remote sensing images and text are aligned, and subsequently, the visual features were fed into an LLM to generate corresponding textual descriptions. Wang et al. \cite{r29} extracted multi-view visual features from fine-tuned visual-language pretrained models and vision-task pretrained models. This method effectively exploits the cross-modal knowledge of pretrained models.
Although recent RSIC methods have demonstrated satisfactory performance, they still struggle to produce accurate and complete captions due to inefficiencies in utilizing textual information within images or in fully integrating information from different modalities and multi-level features.

\section{Methodology}\label{sec3}
\begin{figure*}[tp]
    \centering
    \includegraphics[width=1\textwidth]{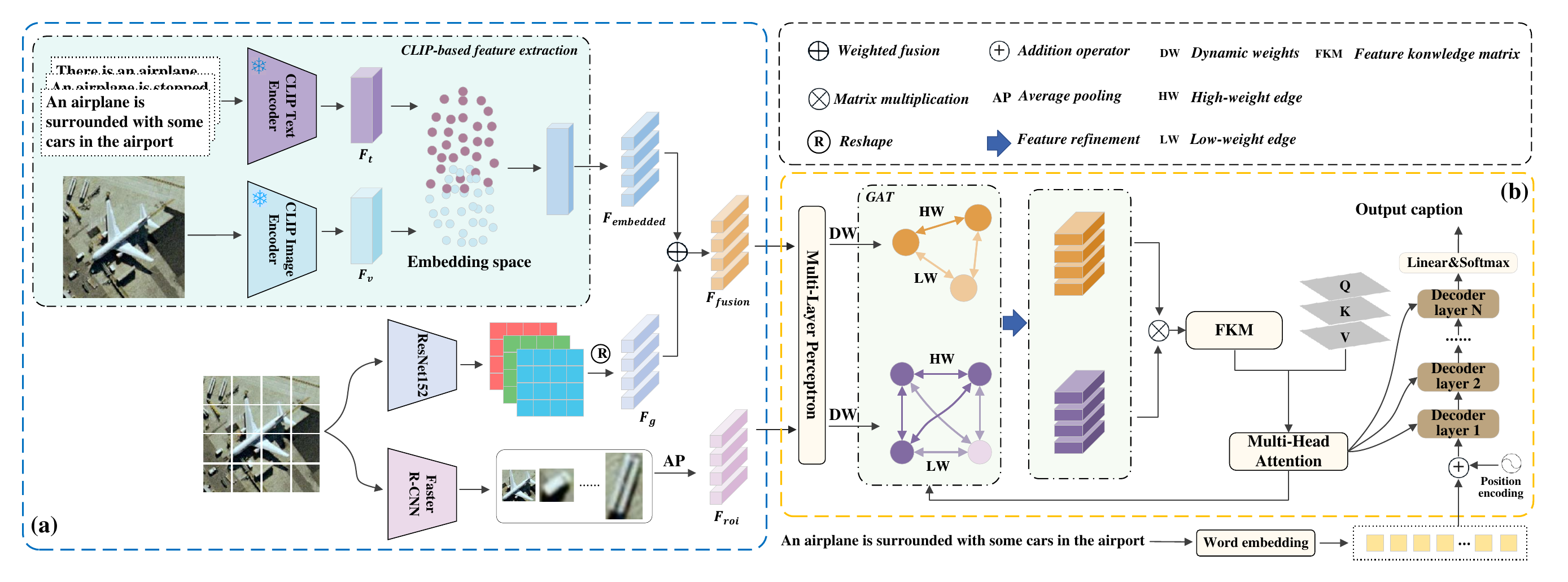}
    \caption{Overall structure of the proposed SFDR. (a) Semantic-spatial feature fusion module, (b) Dynamic graph feature refinement module.}
    \label{fig2}
\end{figure*}

The overall structure of the proposed SFDR is illustrated in Fig. \ref{fig2}, which consists of two main modules: the semantic-spatial feature fusion (SSFF) module and the dynamic graph feature refinement (DGFR) module. In the SSFF module, image and text features are extracted using a pretrained CLIP model and embedded into a shared space to achieve modal complementation, resulting in features enriched with semantic information. These features are then fused with grid-based spatial features to produce a combined representation containing rich semantic and spatial information. In the DGFR module, the ROI features and fused features serve as inputs to a multi-layer perceptron (MLP), which learns from the content of the input features. Through multiple iterations, the MLP progressively identifies the most critical features, generates appropriate dynamic weights, and refines the features by modeling relationships between nodes using a graph attention network. Finally, the decoder generates descriptions based on the refined features.

\subsection{Feature Extraction and Fusion}
The strategy for feature extraction and fusion plays a critical role in the quality of text generation. To optimize this process, we adopt a multi-perspective approach, as illustrated in Fig. \ref{fig2}(a), which involves three key components: (1) extracting visual and textual features using the CLIP model and embedding them into a shared space to obtain semantic features; (2) utilizing the ResNet152 model to extract grid-based spatial features from images and integrating these with the semantic features; and (3) applying the Faster R-CNN model to extract ROI features from images.

\subsubsection{Semantic Feature Extraction} The CLIP model, trained with vision-language alignment, excels at capturing global semantic information relevant to textual descriptions. It focuses on the overall semantic understanding of images, ensuring that the correspondence between images and textual descriptions can be established in a multimodal embedding space. This makes it well-suited for image-text tasks and equips it with remarkable zero-shot capabilities. We utilize the CLIP model to separately extract visual and textual features from the input image $I$, which can be expressed as $\bm{F}_{v} = \mathrm{{CLIP}}(I)$, and $\bm{F}_{t} = \mathrm{CLIP}(T)$, where  $T$ denotes the corresponding text, $\bm{F}_{v} \in \mathbb{R}^{1 \times d_v}$, $\bm{F}_{t} \in \mathbb{R}^{1 \times d_t}$. These features are embedded into a shared space to obtain the semantic information feature, which can be expressed as:
\begin{equation}
    \bm{F}_{\mathrm{embedded}} = \mathrm{Concat}\left( {\bm{F}_{v},~\bm{F}_{t}} \right)
\end{equation}

The resulting embedded feature is a single high-dimensional vector, which can also be represented as $\bm{F}_{\mathrm{embedded}} = \left\{ {f_{1},f_{2},\cdots,f_{d_v + d_t}} \right\}$, where $f_{i} \in \mathbb{R}^{1 \times {({d_v + d_t})}}$.

\subsubsection{Grid-based Spatial Feature Extraction} Grid features emphasize local spatial information, reflecting the geometric structure and background details of an image. They are particularly effective in capturing the spatial layout and contextual background of the scene. Previsous works have exploited CNNs to represent grid features of remote sensing images \cite{r17,n11}. The extracted feature maps by CNNs are split into equal-sized grids with more specific visual patterns compared to CLIP features. Here, we use ResNet152 to extract the grid-based spatial features of the image, which can be expressed as $\bm{F}_{g} = \mathrm{ResNet152}(I)$, where $\bm{F}_{g} \in \mathbb{R}^{H \times W}$, $H$ and $W$ represent the height and width of the feature map, respectively. The expression of semantic information relies on the background and details within the image, as the background and details form a crucial foundation for conveying semantic information. We perform a weighted fusion of semantic features and grid-based spatial features to obtain a composite representation enriched with both semantic and spatial information, which can be expressed as:
\begin{equation}
    \bm{F}_{\mathrm{fusion}} = \alpha \bm{F}_{\mathrm{embedded}} \times H + \left( {1 - \alpha} \right)\bm{F}_{g}
\end{equation}
where $\bm{F}_{\mathrm{fusion}} \in \mathbb{R}^{H \times W}$ and $\alpha$ is the weighting coefficient. Additionally, to ensure dimensional compatibility, the features undergo necessary dimensionality reduction and expansion processes before fusion.

\subsubsection{ROI Feature Extraction} Although CLIP features and grid-based spatial features provide fine-grained information, they lack significant object-level properties. 
ROI features are designed to capture semantically meaningful regions in an image, typically those containing target objects, to aid in identifying objects and regions within the image. By combining ROI features with semantic information, a multi-level feature representation is formed, enabling a more comprehensive understanding of image content. In this work, we employ Faster R-CNN and use a sliding window strategy to extract ROI features from each window region of the image, which can be expressed as $\bm{F}_{\mathrm{roi}} = \mathrm{FRN}(I)$, where $\mathrm{FRN}$ represents the Faster R-CNN model and $\bm{F}_{\mathrm{roi}} \in \mathbb{R}^{{({\frac{({\beta - \gamma})}{\tau} + 1})}^{2} \times l}$ with $\beta$, $\gamma$, $\tau$, $l$ representing the image dimensions, sliding window size, sliding step size, and the length of the feature vector for each ROI, respectively. Notably, we can represent the ROI features as $\bm{F}_{\mathrm{roi}} = \left\{ {\bm{v}_{1},\bm{v}_{2},\cdots,\bm{v}_{k}} \right\}$, where $\bm{v}_{i}$ denotes the feature vector of the $i^{th}$ ROI in the image, which corresponds to the feature representation of the $i^{th}$ object.

\begin{figure}[tp]
    \centering
    \includegraphics[width=1.05\linewidth]{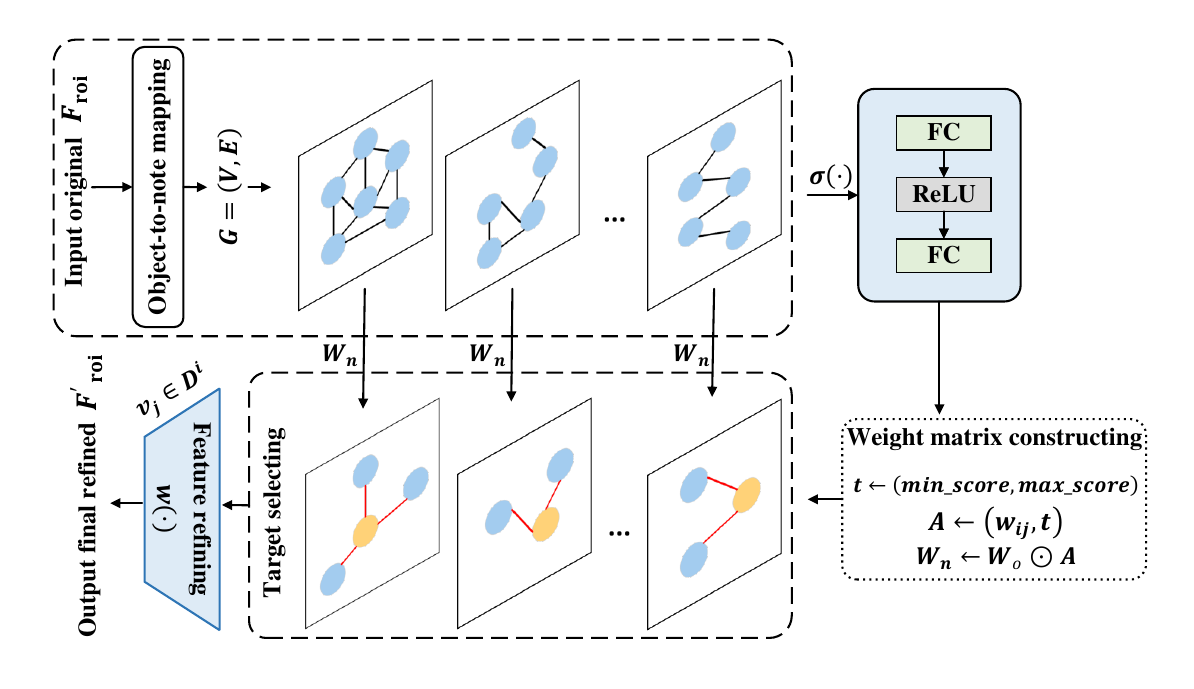}
    \caption{Illustration of the designed DGFR module. In this module, features are treated as nodes, and their relationships are modeled within a graph network. Through weight computation, threshold filtering, and the construction of a weight matrix, feature nodes are reselected. Finally, features are refined based on each node and its neighboring edges.}
    \label{fig3}
\end{figure}

\subsection{Dynamic Graph Feature Refinement}
A graph attention network is utilized to model the relationships between nodes. As illustrated in Fig. \ref{fig3}, using the input ROI features as an example, we construct a graph $\bm{G} = \left( {\bm{V},\bm{E}} \right)$, where $\bm{V} = \left\{ {\bm{v}_{1},\bm{v}_{2},\cdots,\bm{v}_{k}} \right\}$ represents the set of nodes, and $\bm{E} = \left\{ {e_{12},e_{13},\cdots,e_{ij}} \right\}$ represents the set of edges. $e_{ij}$ is the edge between notes $v_{i}$ and $v_{j}$.

Instead of using a static weighting strategy, we employ a multi-layer perceptron, which consists of two full connection layer and ReLU. It dynamically and adaptively adjusts the edge weights based on the interaction between the input features and their context, enhancing the features highly relevant to the scene while suppressing less meaningful ones, which be expressed as:
\begin{equation}
    w_{ij} = \mathrm{MLP}\left( \bm{F}_{\mathrm{roi}} \right) = \sigma\left( {\bm{W}_{2}*\mathrm{ReLU}\left( {\bm{W}_{1}\bm{v}_{i}\bm{v}_{j} + b_{1}} \right) + b_{2}} \right)
\end{equation}
where $\bm{W}_{1} \in \mathbb{R}^{d \times d}$ and $\bm{W}_{2} \in \mathbb{R}^{d \times d}$ are the parameter matrix, $b_{1}$ and $b_{2}$ are the bias term, $\sigma$ is the Sigmoid activation function, and $w_{ij}$ represents the weight corresponding to the edge $e_{ij}$. All the weights $w_{ij}$ form the weight matrix ${W}_{o}$. It is worth mentioning that $b_{1}$ and $b_{2}$ are initialized with small random values to break symmetry and ensure effective gradient propagation. Their values are adjusted based on the optimization of the loss function during training. The optimization of these parameters during training allows the model to refine its feature representations and improve its performance in capturing complex data relationships.



Specifically, to emphasize objects in the ROI features that are most closely related to the scene, a threshold $t$ is utilized to select the edges to retain and construct an adjacency matrix, which can be expressed as:
\begin{equation}
    t = min\_ score + 0.5\left( {max\_ score - min\_ score} \right)
\end{equation}
where $min\_ score$ and $max\_ score$ represent the minimum and maximum attention scores, respectively. Then the adjacency matrix can be established by:
\begin{equation}
\bm{A} = \begin{bmatrix}
a_{11} & \cdots & a_{1j} \\
 \vdots & \ddots & \vdots \\
a_{i1} & \cdots & a_{ij}
\end{bmatrix},
    a_{ij} = \left\{ \begin{matrix}
{1,~~{if~w}_{ij} \geq t} \\
{0,~~otherwise}
\end{matrix} \right.
\end{equation}
where $a_{ij} = 1$ indicates that $w_{ij}$ is a high-weight edge and is retained, while $a_{ij} = 0$ means that $w_{ij}$ is a low-weight edge and is not retained. The updated weight matrix is then computed as $\bm{W}_{n} = \bm{W}_{o}\bigodot \bm{A}$, where $\bigodot$ denotes element-wise multiplication.


After obtaining the edges between a node and its neighboring nodes, the node's features can be updated using the features of its neighboring nodes and the edges. The refined object features are computed based on the node and its neighboring edges, enhancing the ability to model the relationships between the graph nodes. This can be denoted as:
\begin{equation}
    {\bm{\hat{v}}}_{i} = {\sum\limits_{j \in D^{i}}{w_{ij}\bm{v}_{j}\bm{W}_{v}}}
\end{equation}
where $D^{i}$ represents the neighborhood of the node $v_{i}$ in the graph, and $\bm{W}_{v} \in \mathbb{R}^{d \times d}$ is the parameter matrix. Finally, we obtain the refined ROI features $\bm{F}^{'}_\mathrm{roi} = \left\{ {{\bm{\hat{v}}}_{1},{\bm{\hat{v}}}_{2},\cdots{\bm{\hat{v}}}_{m}} \right\} \in \mathbb{R}^{m \times d}$. In the same manner, we can also obtain the refined semantic-spatial fused features $\bm{F}^{'}_\mathrm{fusion} \in \mathbb{R}^{H \times d}$.

The semantic-spatial fused features represent the scene information of the image and text, while the ROI features represent the object information of the image. To establish the connection between the scene and the objects, we use the refined semantic-spatial features and ROI features to compute the relationship between the scene and the objects. This results in a feature knowledge matrix (FKM) $\bm{Z} \in \mathbb{R}^{m \times H}$, which guides the description generation process:
\begin{equation}
    \bm{Z} = \mathrm{Softmax}\left( \frac{\bm{F}^{'}_{roi}\left( \bm{F}^{'}_{fusion} \right)^\mathrm{T}}{\sqrt{d}} \right).
\end{equation}

To establish the correlation between the scene and objects, we use an attention mechanism to compute the Query $\bm{Q}$ and Key $\bm{K}$ using the feature knowledge matrix $\bm{Z}$ and the refined features and apply weighting using the Value $\bm{V}$. This can be expressed as:
\begin{equation}
    \bm{Q} = \bm{F}^{'}_{\mathrm{roi}}\bm{W}^{Q},~~\bm{K} = \bm{F}^{'}_{\mathrm{roi}}(\bm{Z})\bm{W}^{K},~~\bm{V} = \bm{F}^{'}_{\mathrm{roi}}(\bm{Z})\bm{W}^{V}
\end{equation}
\begin{equation}
    \mathrm{Attention}\left( {\bm{Q},\bm{K},\bm{V}} \right) = \mathrm{softmax}\left( \frac{\bm{QK}^\mathrm{T}}{\sqrt{d}} \right)\bm{V}
\end{equation}
where $\bm{W}^{Q} \in \mathbb{R}^{d \times d}$, $\bm{W}^{K} \in \mathbb{R}^{d \times d}$, and $\bm{W}^{V} \in \mathbb{R}^{d \times d}$ are the parameter matrices used to generate the Query, Key, and Value, respectively. $\frac{1}{\sqrt{d}}$ is the scaling factor, which prevents the dot product from becoming too large and ensures gradient stability. The similarity scores calculated are transformed into a probability distribution by the Softmax function. 
 In the multi-head attention mechanism, the attention is performed multiple times, and the results are concatenated and passed through a linear transformation $\bm{W}^{O}$ to obtain the final output:
\begin{equation}
    \mathrm{MHA}\left( {\bm{Q},\bm{K},\bm{V}} \right) = \mathrm{Concat}\left( {{head}_{1},{head}_{2},\cdots,{head}_{h}} \right)\bm{W}^{O}
\end{equation}
\begin{equation}
    {head}_{i} = \mathrm{Attention}\left( {\bm{Q}_{i},\bm{K}_{i},\bm{V}_{i}} \right)
\end{equation}
where $\mathrm{MHA}( \cdot )$ represents the multi-head attention operation and $h$ is the number of heads.

\begin{figure}[tp]
    \centering
    \includegraphics[width=1.05\linewidth]{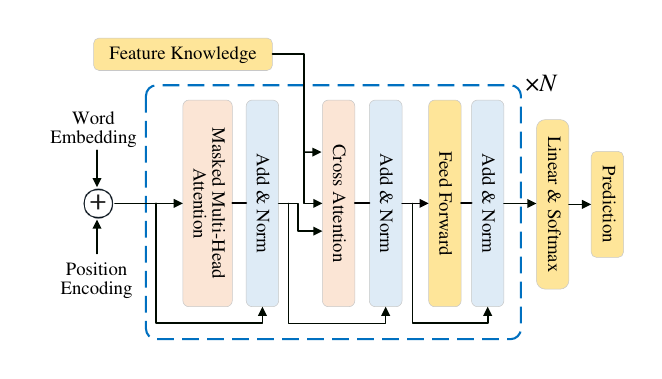}
    \caption{Visualization of the proposed caption generator.}
    \label{fig7}
\end{figure}

\subsection{Caption Generation}
To begin with, we use the Embedding module to extract textual information from raw sentences during training. The output vectors are described as token embeddings $\bm{T_{l}}$, where $l$ is the length of tokens.
\begin{equation}
    \bm{T}_{l} = \left[t_{1},t_{2},\cdots,t_{l}\right].
\end{equation}

Then, a Transformer-based decoder $\mathrm{{Dec}}_{T}$\cite{r36} is employed to generate sentences.
\begin{equation}
    p\left( {\hat{s}}_{t} \middle| {\hat{s}}_{1:t - 1} \right) = \mathrm{Softmax}\left( {\bm{W}^{D}\bm{O}_{N}} \right)
\end{equation}
\begin{equation}
    \bm{O}_{t} = \mathrm{{Dec}}_{T}\left( \bm{Z},\bm{O}_{t - 1} \right),\bm{O}_{1} = \bm{T}_{l}
\end{equation}
where $p\left( {\hat{s}}_{t} \middle| {\hat{s}}_{1:t - 1} \right)$ represents the predicted probability of the $t$-th word conditioned on the token embeddings. Furthermore, $\bm{W}^{D} \in \mathbb{R}^{V_{b} \times d_{h}}$ is a trainable parameter, in which $V_{b}$ is the vocabulary size and $d_{h}$ is the hidden layer size of the decoder. $\bm{O}_{N}$ represents the output of the $N$-th decoding layer. Finally, the generated description is determined based on the probability distribution over the vocabulary.

In addition, to avoid significant changes between training and inference, we apply masked attention in the decoder to ensure that the prediction of each word depends on the preceding generated tokens during inference. Fig. \ref{fig7} provides a visual representation of the decoder. The decoder consists of multiple decoder layers, each of which comprises a masked multi-head attention sub-layer, a cross-attention sub-layer, and a feedforward network.

\subsection{Optimization and Inference Details} 
For description generation, we use the cross-entropy loss to train the model, which can be expressed as:
\begin{equation}
    \mathcal{L}_{CE}(\theta) = - {\sum\limits_{t = 1}^{T}{\log\left( {p_{\theta}\left( {\hat{s}}_{t} \middle| s_{1:t - 1} \right)} \right)}}
\end{equation}
where $\theta$ represents the trainable parameters of the model, $T$ is the length of the sentence, and $s_{1:t - 1}$ denotes the target ground truth (GT) sequence. After training with $\mathcal{L}_{CE}(\theta)$, sequence generation is further optimized through reinforcement learning based on the CIDEr\cite{r40} metric. The gradient expression for optimization is:
\begin{equation}
    \nabla_{\theta}\mathcal{L}_{RL} = - \left( {r\left( \omega_{i}^{S} \right) - r\left( {\hat{\omega}}_{i}^{S} \right)} \right)\nabla_{\theta}{\log\left( {p_{\theta}\omega_{i}^{S}} \right)}
\end{equation}
where $r\left( \omega_{i}^{S} \right)$ and $r\left( {\hat{\omega}}_{i}^{S} \right)$ are the CIDEr rewards for the randomly sampled sentence and the max sampled sentence, respectively.

To ensure consistency between the training and inference processes, two feature input pathways can be designed: an image-text pairs pathway for the training and an image-only input pathway for the inference. During inference, the visual features extracted from the image encoders are fed into the DGFR module. Besides, the model generates sentences conditioned on the feature knowledge matrix and predicts the next tokens one by one under the guidance of the decoder. Furthermore, the beam search method is also applied to boost the diversity of generated captions to some extent.

\section{Experiments}\label{sec4}

\subsection{Datasets}
This study conducted extensive experiments on three benchmark datasets: Sydney-Captions \cite{r6}, UCM-Captions \cite{r6} and RSICD \cite{r14}, to evaluate the ability of the proposed method on image captions across various remote sensing scenarios. Here, we provide a detailed introduction to the structure and characteristics of each dataset:
\begin{enumerate}
    \item Sydney-Captions: This dataset consists of $613$ high-resolution remote sensing images of Sydney and its surrounding areas, with each image having a resolution of $500 \times 500$ pixels. Each image is accompanied by five different textual descriptions. The dataset primarily focuses on describing land-cover distributions in urban environments, including buildings, streets, vegetation, and other man-made structures.
    \item UCM-Captions: This dataset is built upon the UC Merced dataset and covers $21$ different land-cover types, including agricultural areas, forests, residential areas, airports, and more. Each class contains $100$ images with a resolution of $256 \times 256 $ pixels, and each image is accompanied by five unique descriptions.
    \item RSICD: This dataset is a large-scale remote sensing image description dataset containing $10921$ images. All images are sourced from public remote sensing image libraries such as Google Earth, Baidu Map, MapABC, and various commercial remote sensing image providers. Each image has a resolution of $224 \times 224$ pixels and covers more than $30$ scene categories, including buildings, roads, water bodies, farmland, and more. Although each image also has five descriptions, many of the descriptions have only a few variations, leading to higher redundancy.
\end{enumerate}

\begin{table*}[!tp]
\centering
\caption{Comparison Results on the Sydney-Captions Dataset (\%).}
\begin{threeparttable}
\begin{tabular}{ccccccccccc}
\hline
Method          & BLEU-1             & BLEU-2 & BLEU-3 & BLEU-4 & METEOR & ROUGE-L              & CIDEr               & SPICE              & $S_{m}$              & $S_{m}^{*}$             \\ \hline
Soft attention\cite{r14}  & 73.22          & 66.74                  & 62.23                  & 58.20                  & 39.42                 & 71.27          & 249.93          & —              & 104.71          & —               \\
Hard attention\cite{r14}  & 75.91          & 66.10                  & 58.89                  & 52.58                  & 39.98                 & 71.89          & 218.19          & —              & 95.66           & —               \\
MSA\cite{r18}             & 75.07          & 68.00                  & 61.47                  & 55.65                  & 36.74                 & 70.19          & 224.33          & —              & 96.73           & —               \\
M² Transformer\cite{r42}  & 82.25          & 76.19                  & 71.04                  & 66.30                  & 44.20                 & 75.21          & 275.44          & 43.64          & 115.29          & 100.96          \\
Word Sentence\cite{r43}   & 78.91          & 70.94                  & 63.17                  & 56.25                  & 41.81                 & 69.22          & 204.11          & —              & 92.85           & —               \\
GVFGA+LSGA\cite{r32}      & 76.81          & 68.46                  & 61.45                  & 55.04                  & 38.66                 & 70.30          & 245.22          & \textbf{45.32} & 102.31          & 90.91           \\
MLAT\cite{r19}            & 82.96          & 77.77                  & 72.14                  & 67.97                  & 44.24                 & 75.88          & 278.97          & —              & 116.77          & —               \\
RASG\cite{r21}            & 80.00          & 72.20                  & 65.30                  & 59.10                  & 39.10                 & 72.10          & 263.10          & —              & 108.35          & —               \\
Meta-ML\cite{r44}         & 79.58          & 72.74                  & 66.38                  & 60.68                  & 42.47                 & 73.00          & 239.87          & 41.87          & 104.01          & 91.58           \\
Meta-RL\cite{r44}         & 79.75          & 74.21                  & 68.79                  & 63.78                  & 42.79                 & 72.76          & 257.22          & 38.89          & 109.14          & 95.09           \\
Structured attention\cite{r25} & 77.95          & 70.19                  & 63.92                  & 58.61                  & 39.54                 & 72.99          & 237.91          & —          & 102.26          & —          \\
PKG Transformer\cite{r16} & 83.17          & 77.83                  & 72.84                  & 68.24                  & 45.28                 & 77.06          & 284.76          & 44.05          & 118.84          & 103.88          \\
TrTr-CMR\cite{r20}        & 82.70          & 69.94                  & 60.02                  & 51.99                  & 38.03                 & 72.20          & 227.28          & —              & 97.38           & —               \\
RS-CapRet\cite{r45}       & 78.70          & 70.00                  & 62.80                  & 56.40                  & 38.80                 & 70.70          & 239.20          & 43.40          & 101.28          & 89.70           \\ \hline
SFDR(Ours)            & \textbf{85.20} & \textbf{80.18}         & \textbf{75.69}         & \textbf{71.72}         & \textbf{47.34}        & \textbf{79.74} & \textbf{328.64} & 45.22          & \textbf{131.86} & \textbf{114.53} \\ \hline
\end{tabular}
\begin{tablenotes}
        \footnotesize
        \item “—” means the SPICE scores are not reported in the reference papers, so the corresponding $S_{m}^{*}$ score cannot be calculated. The best results are marked in bold.
\end{tablenotes}
\end{threeparttable}
\label{table1}
\end{table*}

\subsection{Evaluation Metrics}
To comprehensively evaluate the quality of generated descriptions, multiple evaluation metrics were employed. These metrics primarily assess the degree of match between the reference descriptions and the generated ones and also measure the accuracy and diversity of the generated text from various perspectives.
\begin{enumerate}
    \item \textit{BLEU-n} \cite{r37}: BLEU (Bilingual Evaluation Understudy) is a classic machine translation evaluation metric used to measure the n-gram matching between the generated descriptions and the reference descriptions. Commonly used versions include BLEU-1, BLEU-2, BLEU-3, and BLEU-4, which evaluate the matching of $1$ to $4$ words (n-grams), respectively.
    \item \textit{METEOR} \cite{r38}: METEOR (Metric for Evaluation of Translation with Explicit Ordering) evaluates the matching between the generated descriptions and reference descriptions at the word level, considering both morphological analysis and semantic matching. Unlike BLEU, METEOR takes into account not only exact word matches but also semantic similarity and morphological variations. The METEOR score is a weighted combination of precision, recall, and F1-score.
    \item \textit{ROUGE-L} \cite{r39}: ROUGLE (Recall-Oriented Understudy for Gisting Evaluation) is a metric primarily used for text summarization. It is based on the longest common subsequence (LCS), which effectively captures word order information. ROUGLE is particularly suitable for evaluating the consistency between the generated and reference descriptions in terms of word order and structure.
    \item \textit{CIDEr} \cite{r40}: CIDEr (Consensus-based Image Description Evaluation) measures the similarity between generated descriptions and reference descriptions based on the matching of TF-IDF weighted n-grams. By focusing on the most distinctive and informative words, CIDEr reflects the alignment of the generated description with the reference descriptions in terms of content and structure. It captures both the word frequency and semantic information effectively, making it highly suitable for evaluating the semantic consistency of descriptions.
    \item \textit{SPICE} \cite{r41}: SPICE (Semantic Propositional Image Caption Evaluation) evaluates based on the matching of semantic graphs, focusing on the expression of objects, relationships, and attributes in the generated descriptions. It helps guide the model in generating more diverse descriptions, reflecting different contents and relationships in the image.
    \item $S_{m}$ and $S_{m}^{*}$: To more intuitively demonstrate the overall performance of the model, we use the standard $S_{m}$ proposed by the 2017 AI Challenger\footnote{https://challenger.ai/competition/caption}, which is the arithmetic mean of the BLEU-4 ($\mathrm{B4}$), METEOR ($\mathrm{M}$), ROUGE-L ($\mathrm{R}$), and CIDEr ($\mathrm{C}$) evaluation metrics. Additionally, to incorporate the SPICE ($\mathrm{S}$) metric, we introduce $S_{m}^{*}$ as shown below: 
\end{enumerate}
\begin{equation}
    S_{m} = \frac{1}{4}\left( \mathrm{{B4 + M + R + C}} \right)
\end{equation}
\begin{equation}
    S_{m}^{*} = \frac{1}{5}\left( \mathrm{{B4 + M + R + C + S}} \right).
\end{equation}

\begin{table*}[!tp]
\centering
\caption{Comparison Results on the UCM-Captions Dataset (\%).}
\begin{threeparttable}
\begin{tabular}{ccccccccccc}
\hline
Method          & BLEU-1             & BLEU-2 & BLEU-3 & BLEU-4 & METEOR & ROUGE-L              & CIDEr               & SPICE              & $S_{m}$              & $S_{m}^{*}$             \\ \hline
Soft attention\cite{r14}  & 74.54                  & 65.45                  & 58.55                  & 52.50                  & 38.86                 & 72.37                 & 261.24                & —                     & 106.24                 & —                       \\
Hard attention\cite{r14}  & 81.57                  & 73.12                  & 67.02                  & 61.82                  & 42.63                 & 76.98                 & 299.47                & —                     & 120.23                 & —                       \\
MSA\cite{r18}             & 83.37                  & 78.22                  & 74.06                  & 70.21                  & 45.04                 & 79.18                 & 325.71                & —                     & 130.04                 & —                       \\
M² Transformer\cite{r42}  & 88.90                  & 85.98                  & 82.74                  & 80.89                  & 51.13                 & 84.45                 & 418.41                & 53.43                 & 158.72                 & 137.66                  \\
Word Sentence\cite{r43}   & 79.31                  & 72.37                  & 66.71                  & 62.02                  & 43.95                 & 71.32                 & 278.71                & —                     & 114.00                 & —                       \\
GVFGA+LSGA\cite{r32}      & 83.19                  & 76.57                  & 71.03                  & 65.96                  & 44.36                 & 78.45                 & 332.70                & 48.53                 & 130.37                 & 114.00                  \\
MLAT\cite{r19}            & 86.42                  & 82.28                  & 78.51                  & 77.98                  & 50.45                 & 83.75                 & 382.50                & —                     & 148.67                 & —                       \\
RASG\cite{r21}            & 85.20                  & 79.30                  & 74.30                  & 69.80                  & 45.70                 & 80.70                 & 338.90                & —                     & 133.78                 & —                       \\
Meta-ML\cite{r44}         & 89.43                  & 85.04                  & 81.37                  & 78.00                  & 52.37                 & 86.87                 & 401.24                & 52.94                 & 154.62                 & 134.28                  \\
Meta-RL\cite{r44}         & 89.51                  & 86.11                  & 83.28                  & 80.71                  & 51.87                 & 85.71                 & 421.28                & 55.72                 & 159.89                 & 139.06                  \\
Structured attention \cite{r25} & 85.38          & 80.35                  & 75.72                  & 71.49                  & 46.32                 & 81.41          & 334.89          & —          & 133.53          & —          \\
PKG Transformer\cite{r16} & 90.48                  & 87.04                  & 84.10                  & 81.39                  & 54.66                 & 86.57                 & 427.49                & 57.01                 & 162.53                 & 141.42                  \\
TrTr-CMR\cite{r20}        & 81.56                  & 70.91                  & 62.20                  & 54.69                  & 39.78                 & 74.42                 & 247.42                & —                     & 104.08                 & —                       \\
RS-CapRet\cite{r45}       & 84.30                  & 77.90                  & 72.20                  & 67.00                  & 47.20                 & 81.70                 & 354.80                & 52.20                 & 137.68                 & 120.58                  \\ \hline
SFDR(Ours)            & \textbf{91.32}         & \textbf{87.96}         & \textbf{85.10}         & \textbf{82.31}         & \textbf{54.76}        & \textbf{88.40}        & \textbf{448.76}       & \textbf{57.58}        & \textbf{168.56}        & \textbf{146.36}         \\ \hline
\end{tabular}
\begin{tablenotes}
        \footnotesize
        \item “—” means the SPICE scores are not reported in the reference papers, so the corresponding $S_{m}^{*}$ score cannot be calculated. The best results are marked in bold.
\end{tablenotes}
\end{threeparttable}
\label{table2}
\end{table*}

\begin{table*}[!tp]
\centering
\caption{Comparison Results on the RSICD (\%).}
\begin{threeparttable}
\begin{tabular}{ccccccccccc}
\hline
Method          & BLEU-1             & BLEU-2 & BLEU-3 & BLEU-4 & METEOR & ROUGE-L              & CIDEr               & SPICE              & $S_{m}$              & $S_{m}^{*}$             \\ \hline
Soft attention\cite{r14}  & 67.53                  & 53.08                  & 43.33                  & 36.17                  & 32.55                 & 61.09                 & 196.43                & —                     & 81.56                  & —                       \\
Hard attention\cite{r14}  & 66.69                  & 51.82                  & 41.64                  & 34.07                  & 32.01                 & 60.84                 & 179.43                & —                     & 76.59                  & —                       \\
MSA\cite{r18}             & 68.69                  & 55.27                  & 46.00                  & 39.21                  & 30.07                 & 56.61                 & 166.76                & —                     & 73.16                  & —                       \\
M² Transformer\cite{r42}  & 68.44                  & 56.57                  & 48.10                  & 41.56                  & 32.69                 & 59.12                 & 258.58                & 45.43                 & 97.99                  & 87.48                   \\
Word Sentence\cite{r43}   & \textbf{72.40}         & 58.61                  & 49.33                  & 42.50                  & 31.97                 & 62.60                 & 206.12                & —                     & 85.80                  & —                       \\
GVFGA+LSGA\cite{r32}      & 67.79                  & 56.00                  & 47.81                  & 41.65                  & 32.85                 & 59.29                 & 260.12                & 46.83                 & 98.46                  & 88.13                   \\
MLAT\cite{r19}            & 68.20                  & 57.18                  & 49.43                  & 42.98                  & 32.22                 & 59.60                 & 265.58                & —                     & 100.10                 & —                       \\
RASG\cite{r21}            & 69.57                  & 57.24                  & 47.10                  & 39.75                  & 34.00                 & \textbf{64.48}        & 247.09                & —                     & 96.33                  & —                       \\
Meta-ML\cite{r44}         & 68.66                  & 56.79                  & 48.39                  & 41.96                  & 32.49                 & 58.82                 & 252.44                & 45.53                 & 96.43                  & 86.25                   \\
Meta-RL\cite{r44}         & 70.56                  & 58.84                  & 50.34                  & 43.69                  & 33.82                 & 60.58                 & 268.37                & 47.60                 & 101.62                 & 90.81                   \\
Structured attention \cite{r25} & 70.16          & 56.14                  & 46.48                  & 39.34                  & 32.91                 & 57.06          & 170.31          & —          & 74.70          & —          \\
PKG Transformer\cite{r16} & 69.67                  & 58.30                  & 50.45                  & 44.31                  & 33.32                 & 60.78                 & 274.01                & 46.91                 & 103.11                 & 91.87                   \\
TrTr-CMR\cite{r20}        & 62.01                  & 39.37                  & 26.71                  & 19.32                  & 23.99                 & 48.95                 & 75.18                 & —                     & 41.86                  & —                       \\
RS-CapRet\cite{r45}       & 72.00                  & \textbf{59.90}         & 50.60                  & 43.30                  & \textbf{37.00}        & 63.30                 & 250.20                & \textbf{47.40}        & 98.45                  & 88.24                   \\ \hline
SFDR(Ours)            & 70.07                  & 58.51                  & \textbf{50.82}         & \textbf{44.81}         & 34.10                 & 61.31                 & \textbf{283.11}       & 46.15                 & \textbf{105.83}        & \textbf{93.90}          \\ \hline
\end{tabular}
\begin{tablenotes}
        \footnotesize
        \item “—” means the SPICE scores are not reported in the reference papers, so the corresponding $S_{m}^{*}$ score cannot be calculated. The best results are marked in bold.
\end{tablenotes}
\end{threeparttable}
\label{table3}
\end{table*}

\subsection{Implementation Details}
In accordance with the comparison methods, each dataset is divided into three subsets: $80\%$ of the image–sentence pairs are used for training, $10\%$  for validation, and 
$10\%$ for testing. However, unlike previous methods, we preserved the original image sizes in all datasets without resizing them.

In the feature extraction stage, we chose ViT-B/32 and Transformer as the image encoder and text encoder of the original CLIP model. To ensure that both image and text are projected into the same feature space for easier similarity calculation, we kept the dimensions of the extracted visual and text features consistent, both being $1 \times 152$. These features were then embedded into the same feature space, resulting in semantic information features with a size of $1 \times 1024$. We use the ResNet152 model pre-trained on the ImageNet dataset to extract the grid-based spatial features of the image. After passing through the final convolutional layer, we obtain a feature map of size $14 \times 14 \times 1024$. This is then reshaped into a $245 \times 2048$ feature map. To combine this with the semantic information features, we first repeat the semantic features $245$ times and then apply a linear transformation to expand their dimensionality to match the grid-based spatial features' size of $245 \times 2048$. Finally, we perform a weighted fusion of both feature sets to obtain the semantic-spatial fused features. We use the Faster R-CNN model in combination with a sliding window strategy to extract ROI features from the images. The settings vary for the three datasets with different image sizes. For the Sydney-Captions dataset, the sliding window size is set to $80 \times 80$ with a stride of $70$; for the UCM-Captions dataset, the window size is $64 \times 64$ with a stride of $32$; for the RSICD dataset, the window size is $32 \times 32$ with a stride of $32$. After applying average pooling and feature flattening, the final ROI feature size is $50 \times 1024$, meaning that $49$ ROI regions are extracted from each image to capture object information comprehensively.

In the training stage, we set the batch size to $64$ and the number of epochs to $40$, with a learning rate of 5e-6. Words that appear more than five times in the training set are retained in the vocabulary. For better performance during inference, we use the beam search algorithm with a beam size of $5$. All experiments are conducted on an NVIDIA GeForce RTX 4090D GPU with PyTorch version 1.10.0.

\subsection{Baseline Methods}
We conducted extensive experimental comparisons with $14$ state-of-the-art methods to demonstrate the effectiveness of the proposed method.
\begin{enumerate}
    \item Soft attention \cite{r14} and Hard attention \cite{r14} are both attention-based methods, where the encoder and decoder use VGG-16 and LSTM, respectively.
    \item MSA \cite{r18} addresses the multi-scale problem by proposing multi-scale attention, which extracts features from different levels and uses a multi-head attention mechanism to obtain contextual features.
    \item M² Transformer \cite{r42}, originally applied to natural images, integrates prior knowledge to learn the relationships between image regions through multi-level representations and develops both low-level and high-level features during the decoding stage.
    \item The Word Sentence framework \cite{r43} consists of a word extractor and a sentence generator. It decomposes RSIC into a word classification task and a word ordering task, aligning more closely with human intuitive understanding.
    \item GVFGA+LSGA \cite{r32} utilize attention gates with global visual features to filter out redundant feature components in the fused image features, providing more prominent image features. It also uses attention gates to filter out irrelevant information in the visual-text fusion features.
    \item MLAT \cite{r19} integrated features from different layers of ResNet50 in the encoder to extract multi-scale information. In the decoder, a Multi-Layer Aggregated Transformer is used to fully leverage the extracted information for sentence generation.
    \item RASG \cite{r21} combines competitive visual features with a recursive attention mechanism, generating better context vectors for the image at each step and enhancing the representation of the current word state.
    \item Meta-ML \cite{r44} and Meta-RL \cite{r44} are multi-task transfer image captioning models based on ResNet and LSTM. ML refers to using meta-learning to train the base model, while RL refers to optimizing the base model using reinforcement learning.
    \item Structured attention \cite{r25} generates pixelwise segmentation masks of semantic contents and the segmentation can be jointly trained with the captioning in a unified framework without requiring any pixelwise annotations.
    \item PKG Transformer \cite{r16} utilizes scene-level and object-level features to provide valuable prior knowledge for RSIC, facilitating the description generation process.
    \item TrTr-CMR \cite{r20} uses a Swin Transformer encoder with a shifted window partitioning scheme for multi-scale visual feature extraction, discovering the intrinsic relationships between objects.
    \item RS-CapRet \cite{r45} is a vision and language method for remote sensing tasks, combining high-performance large decoder language models with image encoders adapted for remote sensing images through language-image pretraining.
\end{enumerate}

\begin{table*}[!tp]
\centering
\caption{Ablation Studies (\%) of Designed Modules on Three Datasets.}
\begin{tabular}{cccccccccccc}
\hline
Dataset                          & Model      & BLEU-1             & BLEU-2             & BLEU-3             & BLEU-4             & METEOR              & ROUGE-L              & CIDEr               & SPICE              & $S_{m}$              & $S_{m}^{*}$             \\ \hline
\multirow{4}{*}{Sydney-Captions} & -SSFF-DGFR & 82.36          & 77.70          & 73.24          & 68.14          & 43.11          & 75.62          & 310.20          & 41.47          & 124.27          & 107.71          \\
                                 & +SSFF-DGFR & 84.43          & 78.78          & 73.74          & 69.02          & 45.99          & 79.48          & 319.96          & 43.48          & 128.61          & 111.59          \\
                                 & -SSFF+DGFR & 83.74          & 79.22          & 74.45          & 70.90          & 45.84          & 77.70          & 316.72          & 44.42          & 127.79          & 111.12          \\
                                 & +SSFF+DGFR & \textbf{85.20} & \textbf{80.18} & \textbf{75.69} & \textbf{71.72} & \textbf{47.34} & \textbf{79.74} & \textbf{328.64} & \textbf{45.22} & \textbf{131.86} & \textbf{114.53} \\ \hline
\multirow{4}{*}{UCM-Captions}    & -SSFF-DGFR & 87.84          & 83.52          & 79.75          & 76.43          & 51.36          & 83.95          & 407.41          & 53.49          & 154.79          & 134.53          \\
                                 & +SSFF-DGFR & 90.60          & 87.03          & 84.15          & 81.52          & 54.08          & 87.31          & 435.05          & 56.70          & 164.49          & 142.93          \\
                                 & -SSFF+DGFR & 90.82          & 87.20          & 83.98          & 81.05          & 54.16          & 87.22          & 422.21          & 56.24          & 161.16          & 140.18          \\
                                 & +SSFF+DGFR & \textbf{91.32} & \textbf{87.96} & \textbf{85.10} & \textbf{82.31} & \textbf{54.76} & \textbf{88.40} & \textbf{448.76} & \textbf{57.58} & \textbf{168.56} & \textbf{146.36} \\ \hline
\multirow{4}{*}{RSICD}           & -SSFF-DGFR & 67.07          & 56.14          & 48.28          & 42.07          & 31.19          & 58.65          & 255.46          & 43.27          & 96.84           & 86.13           \\
                                 & +SSFF-DGFR & 68.11          & 57.12          & 48.88          & 42.50          & 32.41          & 59.63          & 263.87          & 44.46          & 99.60           & 88.57           \\
                                 & -SSFF+DGFR & 68.63          & 57.94          & 50.14          & 43.96          & 32.30          & 60.27          & 268.89          & 44.66          & 101.36          & 90.02           \\
                                 & +SSFF+DGFR & \textbf{70.07} & \textbf{58.51} & \textbf{50.82} & \textbf{44.81} & \textbf{34.10} & \textbf{61.31} & \textbf{283.11} & \textbf{46.15} & \textbf{105.83} & \textbf{93.90}  \\ \hline
\end{tabular}
\label{table4}
\end{table*}

\begin{figure}[tp]
    \centering
    \includegraphics[width=0.5\textwidth]{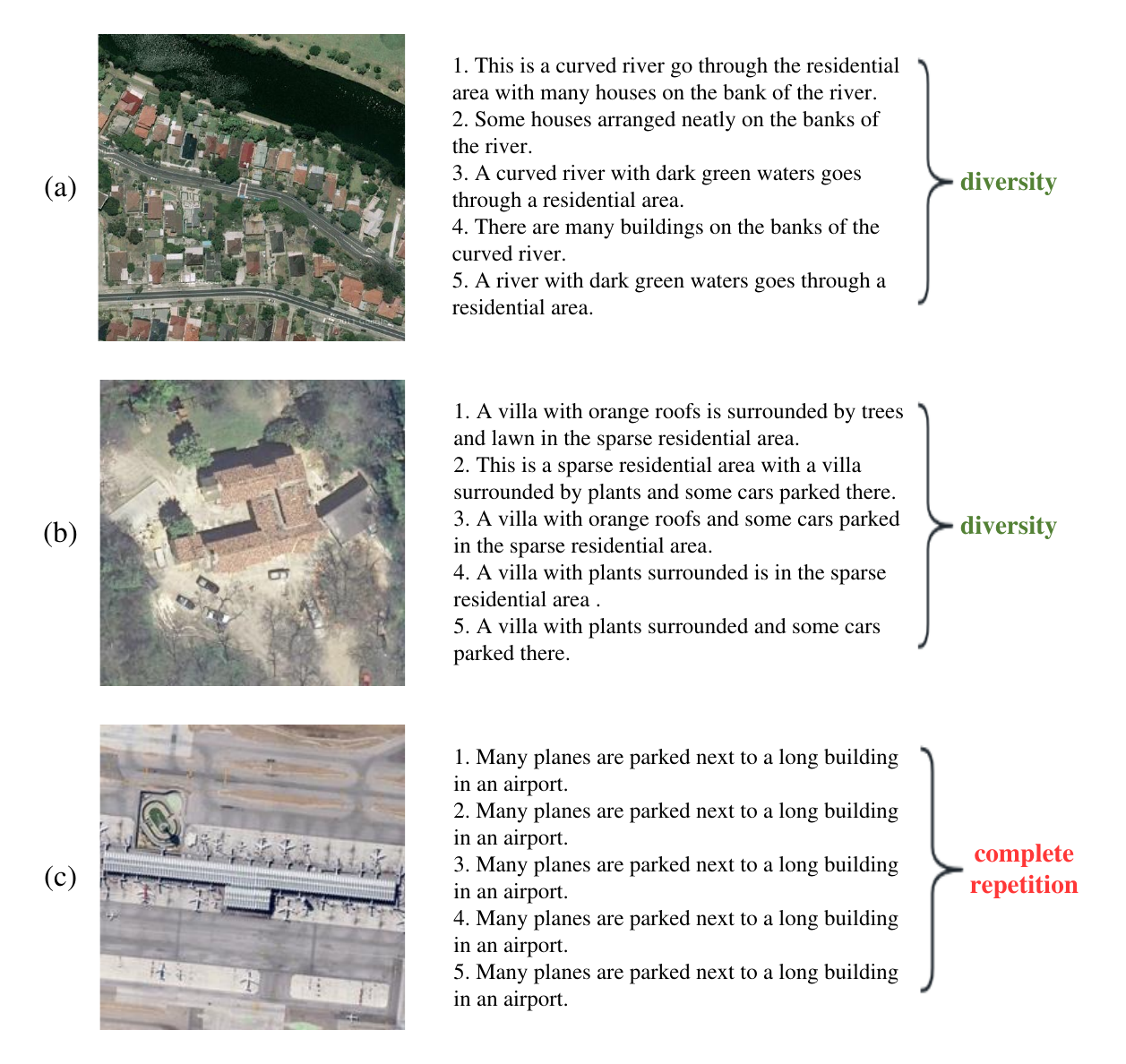}
    \caption{Data samples in the (a) Sydney-Captions dataset, (b) UCM-Captions dataset, and (c) RSICD dataset.}
    \label{fig4}
\end{figure}

\subsection{Quantitative Results}

Table \ref{table1} shows the comparison results of the proposed method against other methods on the Sydney-Captions dataset. Except for the SPICE score, which is marginally lower than that of the GVFGA+LSGA method, our SFDR method outperforms all others across the remaining metrics. Specifically, our SFDR method achieves improvements in BLEU-4 by $5.10\%$, METEOR by $4.55\%$, and ROUGE-L by $3.48\%$ compared to other methods. The most notable improvement is in the CIDEr score, with a significant gain of $15.40\%$, which can be attributed to the mutual enhancement of semantic information between textual and visual data in our SFDR method. The GVFGA+LSGA method achieves the best SPICE score due to its effective attention mechanisms guided by visual features and language states. However, our SFDR method ranks second in SPICE. One possible reason for this is the limited dataset size, which may restrict the model's ability to learn diverse content and relationships, slightly impacting diversity. Additionally, our SFDR method achieves the highest $S_{m}$ and $S_{m}^{*}$ scores, indicating superior overall performance to a certain extent.

\begin{table*}[!tp]
\centering
\caption{Ablation Results (\%) with Different Encoders on Three Datasets.}
\scalebox{0.95}{
\begin{tabular}{cccccccccccccc}
\hline
Dataset                          & CLIP & RN152 & FRN & BLEU-1         & BLEU-2         & BLEU-3         & BLEU-4         & METEOR         & ROUGE-L        & CIDEr           & SPICE          & $S_{m}$              & $S_{m}^{*}$             \\ \hline
\multirow{3}{*}{Sydney-Captions} & \ding{51}    & \ding{55}     & \ding{51}   & 82.76          & 78.43          & 74.61          & 71.33          & 45.60          & 77.51          & 312.39          & 42.46          & 126.71          & 109.86          \\
                                 & \ding{55}    & \ding{51}     & \ding{51}   & 83.74          & 79.22          & 74.45          & 70.90          & 45.84          & 77.70          & 316.72          & 44.42          & 127.79          & 111.12          \\
                                 & \ding{51}    & \ding{51}     & \ding{51}   & \textbf{85.20} & \textbf{80.18} & \textbf{75.69} & \textbf{71.72} & \textbf{47.34} & \textbf{79.74} & \textbf{328.64} & \textbf{45.22} & \textbf{131.86} & \textbf{114.53} \\ \hline
\multirow{3}{*}{UCM-Captions}    & \ding{51}    & \ding{55}     & \ding{51}   & 90.01          & 86.52          & 83.54          & 80.60          & 54.20          & 86.95          & 420.70          & 55.63          & 160.61          & 139.62          \\
                                 & \ding{55}    & \ding{51}     & \ding{51}   & 90.82          & 87.20          & 83.98          & 81.05          & 54.16          & 87.22          & 422.21          & 56.24          & 161.16          & 140.18          \\
                                 & \ding{51}    & \ding{51}     & \ding{51}   & \textbf{91.32} & \textbf{87.96} & \textbf{85.10} & \textbf{82.31} & \textbf{54.76} & \textbf{88.40} & \textbf{448.76} & \textbf{57.58} & \textbf{168.56} & \textbf{146.36} \\ \hline
\multirow{3}{*}{RSICD}           & \ding{51}    & \ding{55}     & \ding{51}   & 68.21          & 57.80          & 50.18          & 43.66          & 32.45          & 59.98          & 266.61          & 45.12          & 100.68          & 89.56           \\
                                 & \ding{55}    & \ding{51}     & \ding{51}   & 68.63          & 57.94          & 50.14          & 43.96          & 32.30          & 60.27          & 268.89          & 44.66          & 101.36          & 90.02           \\
                                 & \ding{51}    & \ding{51}     & \ding{51}   & \textbf{70.07} & \textbf{58.51} & \textbf{50.82} & \textbf{44.81} & \textbf{34.10} & \textbf{61.31} & \textbf{283.11} & \textbf{46.15} & \textbf{105.83} & \textbf{93.90}  \\ \hline
\end{tabular}}
\label{table5}
\end{table*}

\begin{table*}[!tp]
\centering
\caption{Ablation Results (\%) Using Different Grid-based Spatial Features on Three Datasets.}
\scalebox{0.95}{
\begin{tabular}{cccccccccccc}
\hline
Dataset                           & Model             & BLEU-1         & BLEU-2         & BLEU-3         & BLEU-4         & METEOR         & ROUGE-L        & CIDEr           & SPICE          & $S_{m}$              & $S_{m}^{*}$             \\ \hline
\multirow{3}{*}{Sydney-Captions} & RN50+Transformer  & 83.60          & 79.50          & 74.74          & 71.16          & 46.25          & 77.77          & 326.08          & 43.56          & 130.31          & 112.96          \\
                                  & RN101+Transformer & 84.88          & \textbf{80.78} & 75.52          & 71.40          & 47.15          & 78.45          & 326.83          & 44.17          & 131.46          & 114.00          \\
                                  & RN152+Transformer & \textbf{85.20} & 80.18          & \textbf{75.69} & \textbf{71.72} & \textbf{47.34} & 79.74          & \textbf{328.64} & \textbf{45.22} & \textbf{131.86} & \textbf{114.53} \\ \hline
\multirow{3}{*}{UCM-Captions}    & RN50+Transformer  & 89.88          & 86.22          & 83.03          & 80.04          & 53.66          & 87.65          & 431.22          & 56.80          & 163.14          & 141.87          \\
                                  & RN101+Transformer & \textbf{91.36} & \textbf{88.44} & 84.74          & 81.95          & 53.80          & 87.31          & 444.79          & 56.44          & 166.96          & 144.86          \\
                                  & RN152+Transformer & 91.32          & 87.96          & \textbf{85.10} & \textbf{82.31} & \textbf{54.76} & \textbf{88.40} & \textbf{448.76} & \textbf{57.58} & \textbf{168.56} & \textbf{146.36} \\ \hline
\multirow{3}{*}{RSICD}            & RN50+Transformer  & 69.25          & 56.47          & 50.77          & 44.72          & 33.02          & 60.79          & 280.25          & 44.30          & 104.70          & 92.62           \\
                                  & RN101+Transformer & 69.54          & 59.11          & 50.31          & 44.13          & 33.32          & 60.94          & 281.89          & 45.04          & 105.07          & 93.06           \\
                                  & RN152+Transformer & \textbf{70.07} & \textbf{58.51} & \textbf{50.82} & \textbf{44.81} & \textbf{34.10} & \textbf{61.31} & \textbf{283.11} & \textbf{46.15} & \textbf{105.83} & \textbf{93.90}  \\ \hline
\end{tabular}}
\label{table6}
\end{table*}

Table \ref{table2} shows the comparison results of the proposed method with other methods on the UCM-Captions dataset. Notably, our SFDR method achieves the highest scores across all evaluation metrics. Specifically, compared to the best-performing methods, our SFDR method improves BLEU-4 by $1.13\%$, METEOR by $0.18\%$, ROUGE-L by $1.76\%$, CIDEr by $4.98\%$, and SPICE by $1.00\%$. This result is attributed not only to the incorporation of semantic information and multiple RSI representations but also to the precise object information extraction in the images and the effectiveness of feature refinement.

Table \ref{table3} offers the comparison results of the proposed method with other methods on the RSICD dataset. Although the performance on this dataset is not as remarkable as on the other two, our SFDR method still achieves competitive scores in BLEU-3, BLEU-4, and CIDEr, while performing slightly less favorably on the other metrics. We attribute one of the key reasons for this performance discrepancy to the high redundancy in the RSICD dataset, as shown in Fig. \ref{fig4}. Unlike the Sydney-Captions and UCM-Captions datasets, where each image is annotated with five distinct textual descriptions, the RSICD dataset features complete repetition across its five annotations for many images. According to our analysis, $44.44\%$ of the images in RSICD have five identical annotations, $48.93\%$ have $2 - 4$ repeated annotations and only a mere $6.63\%$ of images are paired with five completely distinct descriptions. This limitation results in low-density semantic information and reduced diversity within the dataset, causing the model to favor generating repetitive words and thus leading to suboptimal results. Besides, the RSICD dataset contains some categories of images that belong to very monotonous scenes, such as ``bareland," ``desert," and 'meadow." These scenes have almost no recognizable objects, and for such images, the model often generates only simple descriptions. RS-CapRet, which utilizes large language models to describe remote sensing images, achieves relatively strong performance in BLEU-2, METEOR, and SPICE. Nevertheless, our SFDR method still achieves the best overall performance, as evidenced by the highest $S_{m}$ and $S_{m}^{*}$ scores.

In summary, methods utilizing multi-scale features (e.g., MSA, MLAT, RASG, and TrTr-CMR) offer distinct advantages in processing remote sensing images. Methods incorporating attention guided by visual features (e.g., GVFGA+LSGA, RASG, and PKG Transformer) demonstrate competitive performance. In multitask scenarios, RS-CapRet, Meta-ML, and Meta-RL exhibit significant potential. Comparatively, our method underscores the effectiveness of integrating complementary image-text modality information to enhance semantic understanding and improve object description accuracy through dynamic graph feature refinement.

\begin{figure*}[tp]
    \centering
    \includegraphics[width=1\linewidth]{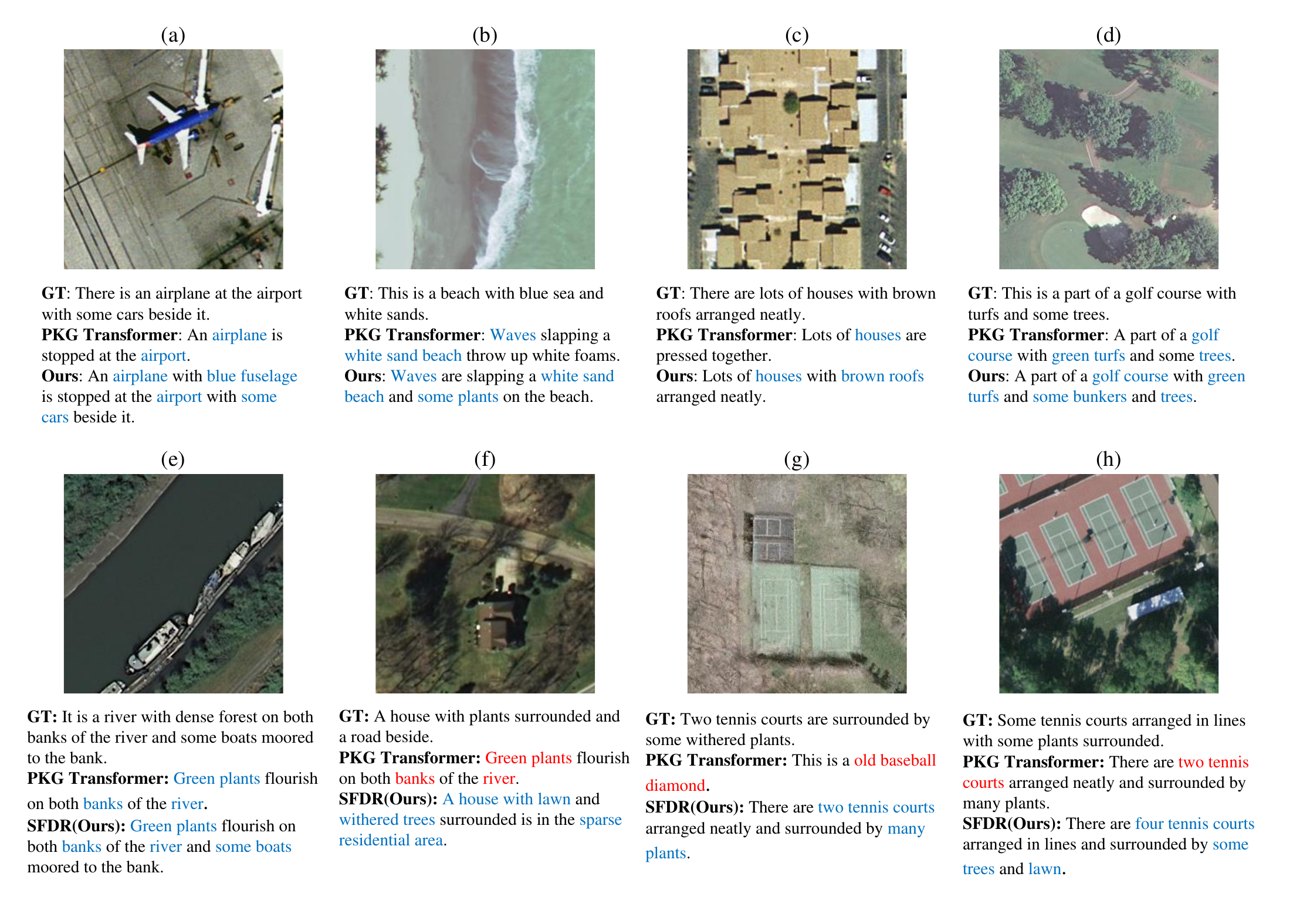}
    \caption{(a)-(h) Typical examples of generated descriptions. The words and phrases highlighted in blue represent accurate descriptions of the image content, while those highlighted in red represent inaccurate or incorrect descriptions of the image contents.}
    \label{fig5}
\end{figure*}

\begin{figure}[htp]
    \centering
    \includegraphics[width=1\linewidth]{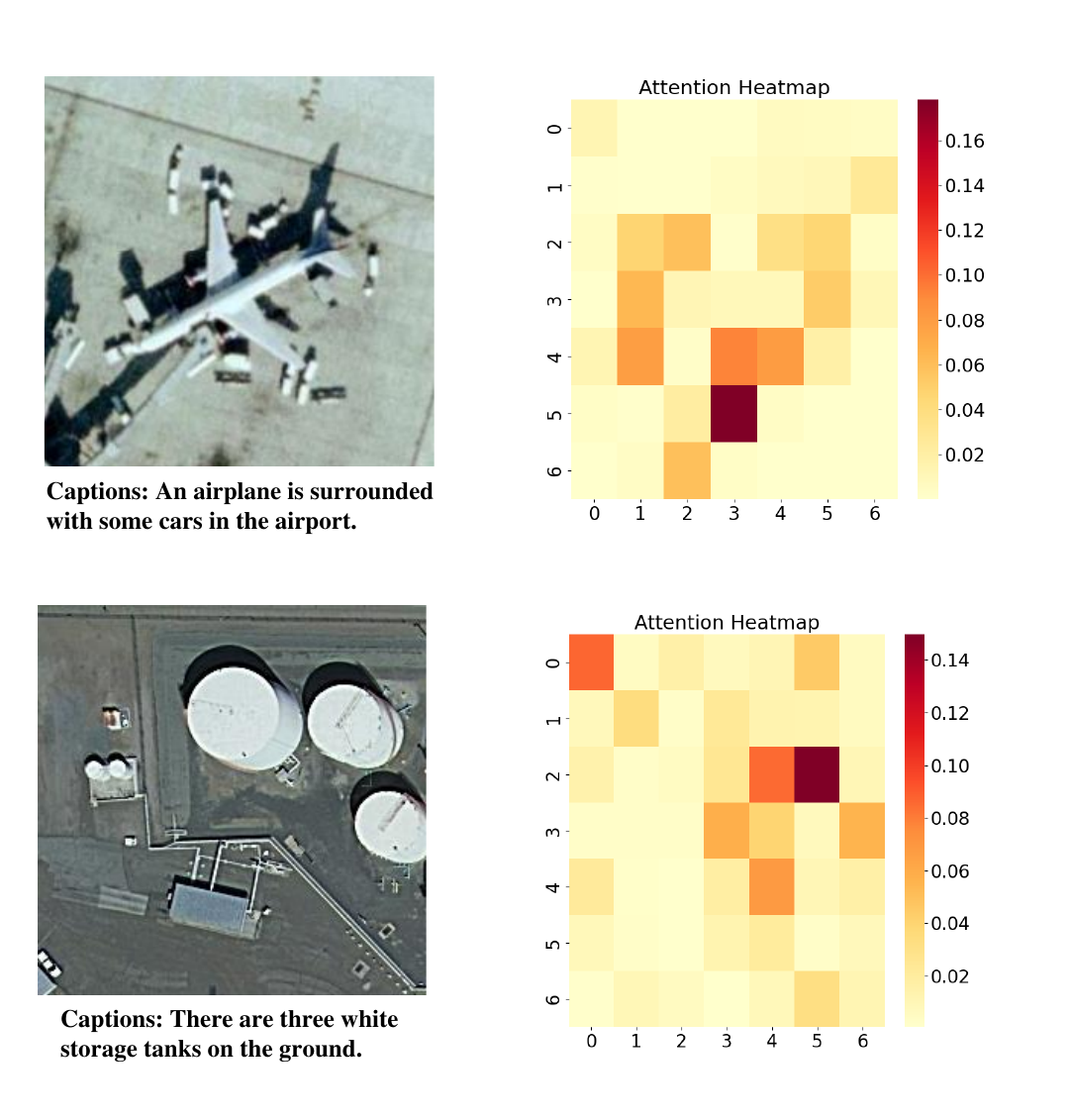}
    \caption{The attention heatmaps related to descriptions.}
    \label{fig6}
\end{figure}

\subsection{Ablation Studies}
In this section, we conduct ablation studies to verify the effectiveness of each designed module in our method. For ``-SSFF”,  no feature fusion is performed, and the encoder consists of ResNet152 and Faster R-CNN. ``-DGFR” denotes the absence of the dynamic weighting mechanism in the graph attention network for feature refinement, the encoders are CLIP, ResNet152, and Faster R-CNN.

The results of the ablation experiments on designed modules on the three datasets are shown in Table \ref{table4}. Overall, removing both the SSFF and DGFR modules simultaneously led to an overall performance decrease of $5.95\%$, $8.08\%$, and $8.27\%$ on the Sydney-Captions, UCM-Captions, and RSICD datasets, respectively. From the results on the Sydney-Captions and UCM-Captions datasets, it can be observed that the performance drop caused by removing the SSFF module (``-SSFF+DGFR") is more pronounced than that caused by removing the DGFR module (``+SSFF-DGFR"). This indicates that the SSFF module plays a more critical role in improving performance on these two datasets. By incorporating semantic information, the model generates richer descriptions that go beyond mere spatial representations. This advantage is closely linked to the textual semantic richness and diversity present in these datasets. Meanwhile, the DGFR module enhances the precise localization of relevant objects through dynamic graph attention for feature refinement. Interestingly, the RSICD dataset reveals a slightly different trend, where the DGFR module's impact is marginally greater than that of the SSFF module. This divergence can be attributed to the previously mentioned data redundancy, which results in less semantic richness and diversity in the image-text pairs of the RSICD dataset compared to the other two datasets. Consequently, the performance improvement contributed by the SSFF module on the RSICD dataset is less pronounced. The combination of both modules (``+SSFF+DGFR") leads to significant performance gains across all metrics, underscoring the effectiveness of integrating these two module designs. Overall, the findings highlight that each module individually enhances model performance, and their joint use maximizes the overall improvements.

Table \ref{table5} shows the ablation experiments on different kinds of encoders. Since the designed DGFR module is primarily used to refine ROI features, all experiments retained Faster R-CNN to ensure the integrity of ROI features while focusing on analyzing the contributions of CLIP and ResNet152 to model performance. The experimental results indicate that removing either CLIP or ResNet152 leads to performance degradation. When using only CLIP, the model performs poorly in fine-grained spatial structure modeling, potentially affecting the understanding of complex scenes. On the other hand, when using only ResNet152, the model lacks semantic alignment knowledge, which may impact the overall coherence of the generated descriptions. When both CLIP and ResNet152 are used together, the model achieves a balance between cross-modal semantic alignment and spatial structure understanding, resulting in the best performance. This result further validates the effectiveness of multi-level feature fusion.

Additionally, considering that many methods use ResNet50 or ResNet101 as the backbone, Table \ref{table6} displays ablation experiments on different grid-based spatial features. The experiments use the same settings to compare the visual features extracted from ResNet50, ResNet101, and ResNet152 on the three datasets. It is worth mentioning that in this paper, ``\textit{B}+Transformer" refers to extracting RSIs visual features from the \textit{B} model and feeding them into a Transformer-based decoder. For example, ``RN50+Transformer" means the Transformer model using the ResNet50 features of RSIs extracted from the pre-trained ResNet50 model. It can be observed that ``RN152+Transformer" outperforms other models on most metric scores. Although several scores in ``RN152+Transformer" do not rank first, like BLEU-1 and BLEU-2, in the UCM-Captions dataset, the difference between its score and the best score is less than $1.00$ point. Overall, compared to ResNet50 and ResNet101, the ResNet152 features may have better performance, so we select it as the grid-based spatial features in the experiments.

\subsection{Qualitative Analysis}
Fig. \ref{fig5} illustrates several examples of image-generated descriptions. Here, ``GT" represents a ground-truth description selected from the original annotations for reference, while ``Ours" corresponds to the descriptions generated by our SFDR method. Descriptions produced by the PKG Transformer method are also included for comparison.

First, our SFDR method effectively identifies prominent objects within the images (e.g., ``airplane," ``houses," and ``trees"), a capability shared by most existing methods. However, our SFDR method goes further by accurately describing objects that are highly relevant but often overlooked by other methods, such as ``cars" in Fig. \ref{fig5}(a), ``bunkers" in Fig. \ref{fig5}(d), and ``boats" in Fig. \ref{fig5}(e). Second, our SFDR method demonstrates a superior ability to describe attribute information related to these objects. For instance, it identifies attributes like ``blue fuselage" in Fig. \ref{fig5}(a) and ``four tennis courts" in Fig. \ref{fig5}(h), offering precise and detailed descriptions of features such as color and quantity.

In contrast, some other methods exhibit errors due to interference from coexisting or visually similar objects. For example, in Fig. \ref{fig5}(g), some methods incorrectly describe the scene as containing a ``baseball diamond," whereas our method correctly identifies ``tennis courts." Moreover, examples like Fig. \ref{fig5}(b) demonstrate our method's ability to not only identify general elements like ``ocean" and ``beach" but also capture finer details such as ``waves." Similarly, Fig. \ref{fig5}(f) includes descriptions of ``lawn" and ``withered trees," showcasing our method's capacity to capture richer scene-related semantic information.

To further illustrate the effectiveness of our method, Fig. \ref{fig6} presents attention heatmaps corresponding to the generated descriptions. These heatmaps visualize how the model focuses on relevant regions of the image during description generation, with darker colors indicating higher attention to specific areas. The images are divided into $7 \times 7$ regions for visualization purposes. For example, when describing ``airplane" and ``cars," the model accurately focuses on the airplane and the scattered vehicles. Similarly, for ``storage tanks", the model correctly identifies both the target objects and their surrounding context. This precise attention alignment ensures that the generated descriptions remain consistent with the visual content, enhancing both their accuracy and relevance.

\begin{figure}[htp]
    \centering
    \includegraphics[width=1\linewidth]{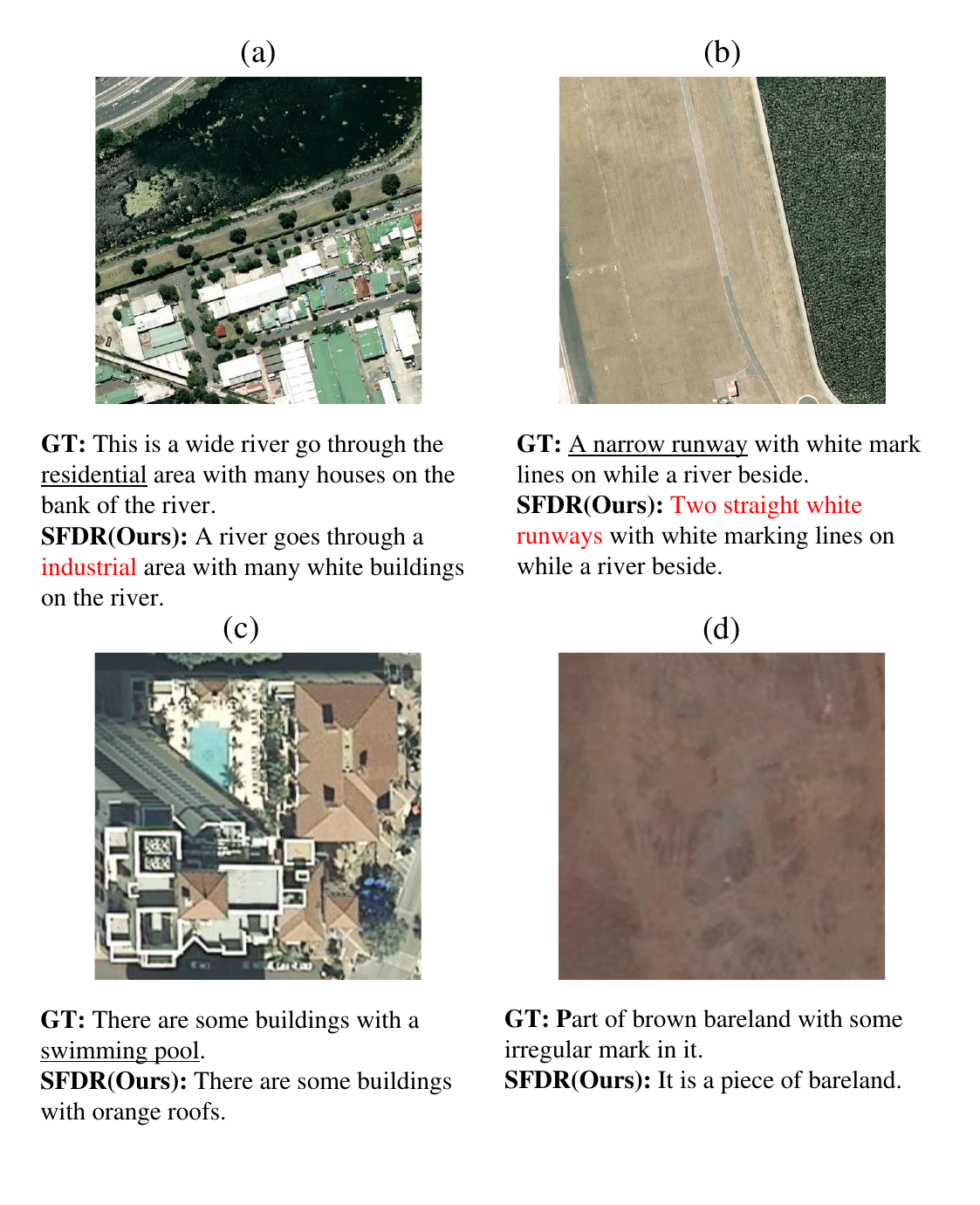}
    \caption{Some unsatisfactory cases generated by the proposed method. The inaccurate or incorrect words are highlighted in red.}
    \label{fig8}
\end{figure}

\subsection{Complexity and Efficiency Analysis}
To comprehensively evaluate the model's complexity and computational efficiency, we report the floating-point operations (FLOPs), number of model parameters, training time, and inference speed (images per second). Table \ref{table7} shows a comparison of different methods on the RSICD. The FLOPs of different methods vary significantly, with RASG achieving the highest at $15.95$G and PKG-Transformer the lowest at $1.58$G. We rank second lowest with $2.05$G, indicating that our method has a relatively low computational complexity. The number of model parameters also varies, with CRSR having the most at $67.26$M and PKG-Transformer the least at $31.94$M. Our model, with $36.55$M parameters, concentrates mainly on the dynamic feature refinement module and the Transformer-based caption generator. In particular, although our method incurs a relatively high training time ($240$ minutes), there is no significant increase in computational complexity, it maintaining a competitive inference speed ($36.40$ images/second), and strikes an effective balance between model complexity with $36.55$ million parameters and computational efficiency and performance.

\subsection{Limitations}
Although our method achieves superior performance, there are still several unsatisfactory results. Fig. \ref{fig8} shows four such cases. For example, in Fig. \ref{fig8}(a), while our method correctly identifies various objects, it mistakenly classifies the residential scene as an extremely similar industrial one. In Fig. \ref{fig8}(b), the riverbank to the right of the runway looks almost identical to the runway, which leads the model to predict it as a runway as well. This indicates that when scenes or objects are difficult for the human eye to distinguish, the model may make incorrect predictions. In Fig. \ref{fig8}(c), the model fails to describe the swimming pool, which might be due to the lack of prior knowledge of this object. Considering the image in Fig. \ref{fig8}(d), which depicts a monotonous scene, although the model's prediction is correct, it tends to generate simpler sentences. This indicates that visually monotonous images like this one limit the model's ability to generate more diverse expressions.

\begin{table}[!tp]
\centering
\captionsetup{justification=centering}
\caption{Computational Efficiency Comparison of Different Methods. All Results are Reported Based on the RSICD.}
\scalebox{0.8}{
\begin{tabular}{ccccc}
\hline
Method              & FLOPs $\downarrow$  & Parameters $\downarrow$ & Training time $\downarrow$ & Inference speed $\uparrow$ \\ \hline
Up-Down\cite{n9}             & 15.89G & 45.18M     & 266min        & 16.73           \\
RASG\cite{r21}                & 15.95G & 53.44M     & 228min        & 16.50           \\
MSISAM\cite{r22}              & —      & —          & 217min        & 16.78           \\
CRSR\cite{r34}                & —      & 67.26M     & 207min        & \textbf{48.53}           \\
Strucured attention\cite{r25} & 5.58G  & 34.60M      & \textbf{175min}        & 1.09            \\
PKG-Transformer\cite{r16}     & \textbf{1.58G}  & \textbf{31.94M}     & 285min        & 37.02           \\ \hline
SFDR(Ours)          & 2.05G  & 36.55M     & 240min        & 36.40           \\ \hline
\end{tabular}}
\label{table7}
\end{table}

\section{Conclusion}\label{sec5}
This paper proposes the SFDR method for the Remote Sensing Image Captioning (RSIC) task. To enhance feature richness and representation across multiple levels, we designed the Semantic-Spatial Feature Fusion (SSFF) module, which explores multi-level feature representations, including CLIP-based features, grid-level features, and Region of Interest (ROI) features. The module integrates advanced feature processing and fusion strategies to provide a more comprehensive representation of remote sensing images. To further improve feature expression, we introduce the Dynamic Graph Feature Refinement (DGFR) module, which refines features using graph-based modeling and dynamic weighting mechanisms. This approach enhances the model's ability to capture intricate relationships and adaptively highlight essential features. 

We conducted comparative experiments against 13 methods to comprehensively evaluate the proposed method. Our SFDR method achieved the highest scores in the metrics BLEU-3, BLEU-4, and CIDEr on three benchmark datasets, surpassing the second-best results by $1.19\%$ to $3.91\%$ in BLEU-3, $1.13\%$ to $5.10\%$ in BLEU-4, and $3.30\%$ to $15.40\%$ in CIDEr. Moreover, our SFDR method also achieved the highest scores in the metrics METEOR and ROUGE-L on Sydney-Captions and UCM-Captions datasets. We have the following findings from the experiments: Through the qualitative analysis, we found our SFDR method can significantly reduce understanding biases by learning more semantic and spatial features. Compared with other methods, the proposed method describes more accurate objects and details. Furthermore, the performance degradation observed in the ablated modules confirms the effectiveness of the designed modules in enhancing the model's performance.

In future work, we plan to extend our research to the remote sensing multimodal event captioning task. This direction offers a promising avenue for exploring more meaningful and practical applications of multimodal analysis in remote sensing.


%



\section*{Acknowledgment}
The authors would like to thank the editors and reviewers for their instructive comments, which helped to improve this article.

\ifCLASSOPTIONcaptionsoff
  \newpage
\fi



%




\bibliographystyle{IEEEtran}
\bibliography{refs}

\begin{thebibliography}{10}
\providecommand{\url}[1]{#1}
\csname url@samestyle\endcsname
\providecommand{\newblock}{\relax}
\providecommand{\bibinfo}[2]{#2}
\providecommand{\BIBentrySTDinterwordspacing}{\spaceskip=0pt\relax}
\providecommand{\BIBentryALTinterwordstretchfactor}{4}
\providecommand{\BIBentryALTinterwordspacing}{\spaceskip=\fontdimen2\font plus
\BIBentryALTinterwordstretchfactor\fontdimen3\font minus \fontdimen4\font\relax}
\providecommand{\BIBforeignlanguage}[2]{{%
\expandafter\ifx\csname l@#1\endcsname\relax
\typeout{** WARNING: IEEEtran.bst: No hyphenation pattern has been}%
\typeout{** loaded for the language `#1'. Using the pattern for}%
\typeout{** the default language instead.}%
\else
\language=\csname l@#1\endcsname
\fi
#2}}
\providecommand{\BIBdecl}{\relax}
\BIBdecl

\bibitem{r1}
M.~Stefanini, M.~Cornia, L.~Baraldi, S.~Cascianelli, G.~Fiameni, and R.~Cucchiara, ``From show to tell: A survey on deep learning-based image captioning,'' \emph{{IEEE} Trans. Pattern Anal. Mach. Intell.}, vol.~45, no.~1, pp. 539--559, Jan. 2023.

\bibitem{r2}
S.~Herdade, A.~Kappeler, K.~Boakye, and J.~Soares, ``Image captioning: Transforming objects into words,'' in \emph{Proc. Adv. Neural Inf. Process. Syst. (NIPS)}, vol.~32, 2019, pp. 11\,135--11\,145.

\bibitem{r3}
W.~Wang, Z.~Chen, and H.~Hu, ``Hierarchical attention network for image captioning,'' in \emph{Proc. AAAI Conf. Artif. Intell. (AAAI)}, vol.~33, no.~01, 2019, pp. 8957--8964.

\bibitem{r4}
S.~J. Rennie, E.~Marcheret, Y.~Mroueh, J.~Ross, and V.~Goel, ``Self-critical sequence training for image captioning,'' in \emph{Proc. IEEE Conf. Comput. Vis. Pattern Recognit. (CVPR)}, Jul. 2017, pp. 7008--7024.

\bibitem{n10}
M.~Liu, B.~Zhou, H.~Hu, C.~Qiu, and X.~Zhang, ``Cross-modal event extraction via visual event grounding and semantic relation filling,'' \emph{Inf. Process. Manag.}, vol.~62, no.~3, p. 104027, 2025.

\bibitem{r5}
O.~Vinyals, A.~Toshev, S.~Bengio, and D.~Erhan, ``Show and tell: A neural image caption generator,'' in \emph{Proc. IEEE Conf. Comput. Vis. Pattern Recognit. (CVPR)}, Jun. 2015, pp. 3156--3164.

\bibitem{r6}
B.~Qu, X.~Li, D.~Tao, and X.~Lu, ``Deep semantic understanding of high resolution remote sensing image,'' in \emph{Proc. Int. Conf. Comput., Inf. Telecommun. Syst. (CITS)}, Jul. 2016, pp. 1--5.

\bibitem{r7}
J.~Lei, X.~Luo, L.~Fang, M.~Wang, and Y.~Gu, ``Region-enhanced convolutional neural network for object detection in remote sensing images,'' \emph{{IEEE} Trans. Geosci. Remote Sens.}, vol.~58, no.~8, pp. 5693--5702, Aug. 2020.

\bibitem{r8}
Y.~Li, Q.~Hou, Z.~Zheng, M.-M. Cheng, J.~Yang, and X.~Li, ``Large selective kernel network for remote sensing object detection,'' in \emph{Proc. IEEE/CVF Int. Conf. Comput. Vis. (ICCV)}, Oct. 2023, pp. 16\,794--16\,805.

\bibitem{r9}
S.~K. Roy, A.~Deria, D.~Hong, B.~Rasti, A.~Plaza, and J.~Chanussot, ``Multimodal fusion transformer for remote sensing image classification,'' \emph{{IEEE} Trans. Geosci. Remote Sens.}, vol.~61, 2023, {Art}. no. 5515620.

\bibitem{r10}
D.~Hong, L.~Gao, N.~Yokoya, J.~Yao, J.~Chanussot, Q.~Du, and B.~Zhang, ``More diverse means better: Multimodal deep learning meets remote-sensing imagery classification,'' \emph{{IEEE} Trans. Geosci. Remote Sens.}, vol.~59, no.~5, pp. 4340--4354, May 2020.

\bibitem{r11}
X.~Zhang, W.~Wu, M.~Zhang, W.~Yu, and P.~Ghamisi, ``Prototypical unknown-aware multiview consistency learning for open-set cross-domain remote sensing image classification,'' \emph{{IEEE} Trans. Geosci. Remote Sens.}, vol.~62, 2024, {Art}. no. 5643616.

\bibitem{n7}
L.~Bai, Q.~Liu, C.~Li, Z.~Ye, M.~Hui, and X.~Jia, ``Remote sensing image scene classification using multiscale feature fusion covariance network with octave convolution,'' \emph{{IEEE} Trans. Geosci. Remote Sens.}, vol.~60, 2022, {Art}. no. 5620214.

\bibitem{n8}
Z.~Ye, Y.~Zhang, J.~Zhang, W.~Li, and L.~Bai, ``A multiscale incremental learning network for remote sensing scene classification,'' \emph{{IEEE} Trans. Geosci. Remote Sens.}, vol.~62, 2024, {Art}. no. 5606015.

\bibitem{r12}
Z.~Shi and Z.~Zou, ``Can a machine generate humanlike language descriptions for a remote sensing image?'' \emph{{IEEE} Trans. Geosci. Remote Sens.}, vol.~55, no.~6, pp. 3623--3634, Jun. 2017.

\bibitem{r13}
B.~Wang, X.~Lu, X.~Zheng, and X.~Li, ``Semantic descriptions of high-resolution remote sensing images,'' \emph{{IEEE} Geosci. Remote Sens. Lett.}, vol.~16, no.~8, pp. 1274--1278, Aug. 2019.

\bibitem{r14}
X.~Lu, B.~Wang, X.~Zheng, and X.~Li, ``Exploring models and data for remote sensing image caption generation,'' \emph{{IEEE} Trans. Geosci. Remote Sens.}, vol.~56, no.~4, pp. 2183--2195, Apr. 2018.

\bibitem{r15}
Y.~Li, S.~Fang, L.~Jiao, R.~Liu, and R.~Shang, ``A multi-level attention model for remote sensing image captions,'' \emph{Remote Sens.}, vol.~12, no.~6, p. 939, Mar. 2020.

\bibitem{r16}
L.~Meng, J.~Wang, Y.~Yang, and L.~Xiao, ``Prior knowledge-guided transformer for remote sensing image captioning,'' \emph{{IEEE} Trans. Geosci. Remote Sens.}, vol.~61, 2023, {Art}. no. 4706213.

\bibitem{r17}
K.~Zhao and W.~Xiong, ``Cooperative connection transformer for remote sensing image captioning,'' \emph{{IEEE} Trans. Geosci. Remote Sens.}, vol.~62, 2024, {Art}. no. 5607314.

\bibitem{r18}
X.~Ma, R.~Zhao, and Z.~Shi, ``Multiscale methods for optical remote-sensing image captioning,'' \emph{{IEEE} Geosci. Remote Sens. Lett.}, vol.~18, no.~11, pp. 2001--2005, Nov. 2021.

\bibitem{r19}
C.~Liu, R.~Zhao, and Z.~Shi, ``Remote-sensing image captioning based on multilayer aggregated transformer,'' \emph{{IEEE} Geosci. Remote Sens. Lett.}, vol.~19, 2022, {Art}. no. 6506605.

\bibitem{r20}
Y.~Wu, L.~Li, L.~Jiao, F.~Liu, X.~Liu, and S.~Yang, ``{TrTr-CMR}: Cross-modal reasoning dual transformer for remote sensing image captioning,'' \emph{{IEEE} Trans. Geosci. Remote Sens.}, vol.~62, 2024, {Art}. no. 5643912.

\bibitem{r21}
Y.~Li, X.~Zhang, J.~Gu, C.~Li, X.~Wang, X.~Tang, and L.~Jiao, ``Recurrent attention and semantic gate for remote sensing image captioning,'' \emph{{IEEE} Trans. Geosci. Remote Sens.}, vol.~60, 2021, {Art}. no. 5608816.

\bibitem{r22}
X.~Zhang, Y.~Li, X.~Wang, F.~Liu, Z.~Wu, X.~Cheng, and L.~Jiao, ``Multi-source interactive stair attention for remote sensing image captioning,'' \emph{Remote Sens.}, vol.~15, no.~3, p. 579, Jan. 2023.

\bibitem{r23}
W.~Huang, Q.~Wang, and X.~Li, ``Denoising-based multiscale feature fusion for remote sensing image captioning,'' \emph{{IEEE} Geosci. Remote Sens. Lett.}, vol.~18, no.~3, pp. 436--440, Mar. 2021.

\bibitem{r24}
Z.~Yuan, X.~Li, and Q.~Wang, ``Exploring multi-level attention and semantic relationship for remote sensing image captioning,'' \emph{IEEE Access}, vol.~8, pp. 2608--2620, Dec. 2019.

\bibitem{r25}
R.~Zhao, Z.~Shi, and Z.~Zou, ``High-resolution remote sensing image captioning based on structured attention,'' \emph{{IEEE} Trans. Geosci. Remote Sens.}, vol.~60, 2021, {Art}. no. 5603814.

\bibitem{n3}
Z.~Yang, Y.~Lu, J.~Wang, X.~Yin, D.~Florencio, L.~Wang, C.~Zhang, L.~Zhang, and J.~Luo, ``{TAP}: Text-aware pre-training for text-vqa and text-caption,'' in \emph{Proc. IEEE/CVF Conf. Comput. Vis. Pattern Recognit. (CVPR)}, Jun. 2021, pp. 8751--8761.

\bibitem{n4}
P.~Zhang, X.~Li, X.~Hu, J.~Yang, L.~Zhang, L.~Wang, Y.~Choi, and J.~Gao, ``{VinVL}: Revisiting visual representations in vision-language models,'' in \emph{Proc. IEEE/CVF Conf. Comput. Vis. Pattern Recognit. (CVPR)}, Jun. 2021, pp. 5579--5588.

\bibitem{n5}
J.~Chen, H.~Guo, K.~Yi, B.~Li, and M.~Elhoseiny, ``{VisualGPT}: Data-efficient adaptation of pretrained language models for image captioning,'' in \emph{Proc. IEEE/CVF Conf. Comput. Vis. Pattern Recognit. (CVPR)}, Jun. 2022, pp. 18\,030--18\,040.

\bibitem{n6}
W.~Kim, B.~Son, and I.~Kim, ``Vilt: Vision-and-language transformer without convolution or region supervision,'' in \emph{Proc. 38th Int. Conf. Mach. Learn. (ICML)}, 2021, pp. 5583--5594.

\bibitem{r26}
A.~Radford, J.~W. Kim, C.~Hallacy, A.~Ramesh, G.~Goh, S.~Agarwal, G.~Sastry, A.~Askell, P.~Mishkin, J.~Clark \emph{et~al.}, ``Learning transferable visual models from natural language supervision,'' in \emph{Proc. 38th Int. Conf. Mach. Learn. (ICML)}, 2021, pp. 8748--8763.

\bibitem{r29}
S.~Wang, Q.~Lin, X.~Ye, Y.~Liao, D.~Quan, Z.~Jin, B.~Hou, and L.~Jiao, ``Multi-view feature fusion and visual prompt for remote sensing image captioning,'' \emph{{IEEE} Trans. Geosci. Remote Sens.}, vol.~62, 2024, {Art}. no. 4708217.

\bibitem{r28}
C.~Yang, Z.~Li, and L.~Zhang, ``Bootstrapping interactive image-text alignment for remote sensing image captioning,'' \emph{{IEEE} Trans. Geosci. Remote Sens.}, vol.~62, 2024, {Art}. no. 5607512.

\bibitem{n2}
L.~Meng, J.~Wang, R.~Meng, Y.~Yang, and L.~Xiao, ``A multiscale grouping transformer with clip latents for remote sensing image captioning,'' \emph{{IEEE} Trans. Geosci. Remote Sens.}, vol.~62, 2024, {Art}. no. 4703515.

\bibitem{r30}
X.~Zhang, X.~Wang, X.~Tang, H.~Zhou, and C.~Li, ``Description generation for remote sensing images using attribute attention mechanism,'' \emph{Remote Sens.}, vol.~11, no.~6, p. 612, Mar. 2019.

\bibitem{r31}
Z.~Zhang, W.~Diao, W.~Zhang, M.~Yan, X.~Gao, and X.~Sun, ``{LAM}: Remote sensing image captioning with label-attention mechanism,'' \emph{Remote Sens.}, vol.~11, no.~20, p. 2349, Oct. 2019.

\bibitem{r32}
Z.~Zhang, W.~Zhang, M.~Yan, X.~Gao, K.~Fu, and X.~Sun, ``Global visual feature and linguistic state guided attention for remote sensing image captioning,'' \emph{{IEEE} Trans. Geosci. Remote Sens.}, vol.~60, 2021, {Art}. no. 5615216.

\bibitem{r35}
Y.~Li, X.~Zhang, X.~Cheng, X.~Tang, and L.~Jiao, ``Learning consensus-aware semantic knowledge for remote sensing image captioning,'' \emph{Pattern Recognit.}, vol. 145, p. 109893, Jan. 2024.

\bibitem{r33}
Z.~Yang, Q.~Li, Y.~Yuan, and Q.~Wang, ``{HCNet}: Hierarchical feature aggregation and cross-modal feature alignment for remote sensing image captioning,'' \emph{{IEEE} Trans. Geosci. Remote Sens.}, vol.~62, 2024, {Art}. no. 5624711.

\bibitem{r34}
Z.~Li, W.~Zhao, X.~Du, G.~Zhou, and S.~Zhang, ``Cross-modal retrieval and semantic refinement for remote sensing image captioning,'' \emph{Remote Sens.}, vol.~16, no.~1, p. 196, Jan. 2024.

\bibitem{n1}
H.~Wei and Z.~Chen, ``Improving generalization of image captioning with unsupervised prompt learning,'' \emph{arXiv:2308.02862}, 2023.

\bibitem{n11}
K.~Zhao and W.~Xiong, ``Exploring region features in remote sensing image captioning,'' \emph{Int. J. Appl. Earth Obs. Geoinf.}, vol. 127, p. 103672, 2024.

\bibitem{r36}
A.~Vaswani, ``Attention is all you need,'' in \emph{Proc. Adv. Neural Inf. Process. Syst. (NIPS)}, 2017, pp. 5998--6008.

\bibitem{r40}
R.~Vedantam, C.~Lawrence~Zitnick, and D.~Parikh, ``{CIDEr}: Consensus-based image description evaluation,'' in \emph{Proc. IEEE Conf. Comput. Vis. Pattern Recognit. (CVPR)}, Jun. 2015, pp. 4566--4575.

\bibitem{r42}
M.~Cornia, M.~Stefanini, L.~Baraldi, and R.~Cucchiara, ``Meshed-memory transformer for image captioning,'' in \emph{Proc. IEEE/CVF Conf. Comput. Vis. Pattern Recognit. (CVPR)}, Jun. 2020, pp. 10\,578--10\,587.

\bibitem{r43}
Q.~Wang, W.~Huang, X.~Zhang, and X.~Li, ``Word--sentence framework for remote sensing image captioning,'' \emph{{IEEE} Trans. Geosci. Remote Sens.}, vol.~59, no.~12, pp. 10\,532--10\,543, Dec. 2020.

\bibitem{r44}
Q.~Yang, Z.~Ni, and P.~Ren, ``Meta captioning: A meta learning based remote sensing image captioning framework,'' \emph{ISPRS J. Photogramm. Remote Sens.}, vol. 186, pp. 190--200, Apr. 2022.

\bibitem{r45}
J.~D. Silva, J.~Magalh{\~a}es, D.~Tuia, and B.~Martins, ``Large language models for captioning and retrieving remote sensing images,'' \emph{arXiv:2402.06475}, 2024.

\bibitem{r37}
K.~Papineni, S.~Roukos, T.~Ward, and W.-J. Zhu, ``{BLEU}: a method for automatic evaluation of machine translation,'' in \emph{Proc. 40th Annu. Meeting Assoc. Comput. Linguistics}, 2002, pp. 311--318.

\bibitem{r38}
S.~Banerjee and A.~Lavie, ``{METEOR}: An automatic metric for mt evaluation with improved correlation with human judgments,'' in \emph{Proc. ACL Workshop Intrinsic Extrinsic Eval. Measures Mach. Transl. Summarization}, 2005, pp. 65--72.

\bibitem{r39}
C.-Y. Lin, ``{ROUGE}: A package for automatic evaluation of summaries,'' in \emph{Proc. Text Summarization Branches Out}, 2004, pp. 74--81.

\bibitem{r41}
P.~Anderson, B.~Fernando, M.~Johnson, and S.~Gould, ``{SPICE}: Semantic propositional image caption evaluation,'' in \emph{Proc. Eur. Conf. Comput. Vis. (ECCV)}, Oct. 2016, pp. 382--398.

\bibitem{n9}
P.~Anderson, X.~He, C.~Buehler, D.~Teney, M.~Johnson, S.~Gould, and L.~Zhang, ``Bottom-up and top-down attention for image captioning and visual question answering,'' in \emph{Proc. IEEE Conf. Comput. Vis. Pattern Recognit. (CVPR)}, Jun. 2018, pp. 6077--6086.

\end{thebibliography}

%








\end{document}